\documentclass{article}
\usepackage{ijcai26}

\usepackage{times}
\usepackage{soul}
\usepackage{url}
\usepackage[hidelinks]{hyperref}
\usepackage[utf8]{inputenc}
\usepackage[small]{caption}
\usepackage{graphicx}
\usepackage{amsmath}
\usepackage{amsthm}
\usepackage{booktabs}
\usepackage{algorithm}
\usepackage{algorithmic}
\usepackage[switch]{lineno}

\usepackage{amsmath}
\usepackage{amssymb} 
\usepackage{amsthm}
\usepackage{booktabs}
\usepackage{makecell}
\usepackage{multirow}
\usepackage{mathtools}
\usepackage{multicol}
\usepackage{pifont} 
\usepackage{bm}

\newtheorem{theorem}{Theorem}

\title{Transport-Coupled Bayesian Flows for Molecular Graph Generation}

\author{
Yida Xiong$^{1,2}$\and
Jiameng Chen$^1$\and
Kun Li$^1$\and
Hongzhi Zhang$^1$\and \\
Xiantao Cai$^1$\and
Lei Lei$^{3,*}$\and
Wenbin Hu$^{1,2,4,*}$\and
\\
\affiliations
$^1$ School of Computer Science, Wuhan University, Wuhan, China\\
$^2$ Hubei Jiangxia Laboratory, China\\
$^3$ Institute of Information on Traditional Chinese Medicine, China Academy of Chinese Medical Sciences, China\\
$^4$ Wuhan University Shenzhen Research Institute, Shenzhen, China\\
$^*$ Corresponding author\\
\emails
\{yidaxiong, jiameng.chen, likun98, zhanghongzhi, caixiantao, hwb\}@whu.edu.cn,
leilei@mail.cintcm.ac.cn
}

\begin{document}

\maketitle

\vspace{-20pt}


\begin{abstract} 
    Molecular graph generation (MGG) is essentially a multi-class generative task, aimed at predicting categories of atoms and bonds under strict chemical and structural constraints. However, many prevailing diffusion paradigms learn to regress numerical embeddings and rely on a hard discretization rule during sampling to recover discrete labels. This introduces a fundamental discrepancy between training and sampling. While models are trained for point-wise numerical fidelity, the sampling process fundamentally relies on crossing categorical decision boundaries. This discrepancy forces the model to expend efforts on intra-class variations that become irrelevant after discretization, ultimately compromising diversity, structural statistics, and generalization performance. Therefore, we propose \textbf{TopBF}, a unified framework that (i) performs MGG directly in continuous parameter distributions, (ii) learns graph-topological understanding through a Quasi-Wasserstein optimal-transport coupling under geodesic costs, and (iii) supports controllable, property-conditioned generation during sampling without retraining the base model. TopBF innovatively employs cumulative distribution function (CDF) to compute category probabilities induced by the Gaussian channel, thereby unifying the training objective with the sampling discretization operation. Experiments on QM9 and ZINC250k demonstrate superior structural fidelity and efficient generation with improved performance. The reproducible code, datasets, and formal theoretical proofs are provided in the Supplementary Material.
\end{abstract}

\section{Introduction}

\begin{figure}
    \centering
    \includegraphics[width=1.\linewidth]{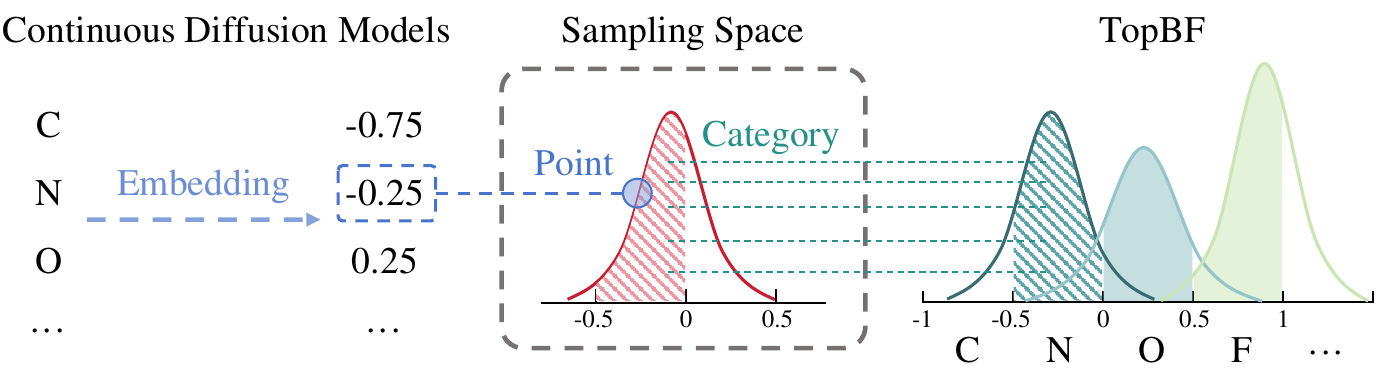}
    \caption{Continuous diffusion models are agnostic of the discretization process and only pursue numerical fitting during training, while TopBF directly calculates the probability of each category.}
    \label{fig:motivation}
\end{figure}


Molecular graph generation (MGG) is a critical yet challenging endeavor in modern computational drug discovery \cite{li2025graph}, facilitating the advancement of downstream tasks such as molecular optimization \cite{zhu2023sample,XIONG2026103907} and drug–target binding affinity prediction \cite{wu2024psc,li2025contrastive}. Early computational approaches utilizing deep learning attempted to generate molecular graphs in a one-shot manner \cite{zang2020moflow,kong2022molecule}. However, these methods often struggle to model complex structures effectively. In recent years, diffusion models have demonstrated remarkable performance across various generation tasks \cite{gong2024text,chen2025antibody}. These models are generally divided into discrete and continuous types and have been progressively integrated into molecular graph generation.

Naturally suited for discrete graph structures, discrete diffusion models typically define a forward Markov process that corrupts the graph structure through discrete edits. Conversely, the reverse process learns to restore the graph by framing generation as an iterative classification task \cite{vignac2023digress,xu2024discrete}. 
However, these discrete diffusion models sacrifice the smoother optimization and dynamic generation afforded by continuous representations. Continuous diffusion models have also demonstrated success in this domain \cite{jo2022score,luo2023fast}. They operate on a continuous representation of a graph, involving a forward noising process that incrementally perturbs this representation with continuous noise while training a core network to reverse this corruption.
The primary objective of continuous diffusion models is to predict embedded representations in continuous space and train a neural network to regress those values under injected noise. However, the final output must be discretized by rounding, which is not a benign post-processing step. As shown in Figure~\ref{fig:motivation}, it is evident that throughout the entire training phase, these models remain agnostic regarding the discretization rule. The resulting mismatch between training and sampling can lead to inefficient learning and brittle generation. In particular, point-wise regression errors near category boundaries may be penalized mildly during training but induce catastrophic category flips at inference, which is especially problematic for generation.

Recently, Bayesian Flow Networks (BFNs) \cite{graves2023bayesian,song2024unified} provide an alternative paradigm for discrete generation by evolving distribution parameters through a continuous-time Bayesian updating process. Instead of directly predicting discrete values, BFNs propagate a parameter state that represents uncertainty and update it via pseudo-observations, yielding a continuous and differentiable learning process for discrete data. 
However, BFNs are not topology-aware for molecular graphs, since they adopt independent Gaussian distributions between variables to retain closed-form Bayesian updates. As a result, BFNs are mainly rewarded for getting each position right in isolation, rather than for capturing how labels should cohere with the overall graph structure.

To address the aforementioned issues, we introduce \textbf{TopBF}, a molecular graph generative BFN that performs generation in continuous distribution-parameter space and uses a CDF-based decoding to compute category probabilities from continuous parameters, thereby aligning the training objective with the same discretization rule used during sampling. Furthermore, we make TopBF topology-aware by introducing a Quasi-Wasserstein transport coupling defined on graph geodesic costs, which compares predicted categorical mass on atoms and bonds with the ground-truth structure. This geometry-respecting discrepancy encourages local, structure-consistent adjustments and discourages topology-inconsistent long-range rearrangements. TopBF also achieves conditional generation via property-error gradient (PEG) guidance.
Notably, TopBF achieves superior performance with the minimum sampling steps, accelerating molecular graph generation in practice.
We summarize our key contributions as follows:

\begin{itemize}
    \item We observe the mismatch defect in continuous diffusion methods and propose TopBF that unifies continuous-parameter training with discrete sampling for MGG.
    \item We inject topology awareness into TopBF training via a Quasi-Wasserstein transport coupling with graph-geodesic ground costs by aligning predicted categorical mass with the molecular structure in a geometry-consistent manner.
    \item TopBF supports training-free and property-conditioned generation via guided sampling while requiring the minimum sampling steps, achieving both effective control and efficient MGG.
\end{itemize}

\section{Related Work}

Deep generative models for MGG are typically categorized into autoregressive and one-shot models. The former generate graphs incrementally, adding atoms and bonds in an autoregressive manner \cite{you2018graph,you2018graphrnn,jin2018junction,jin2020hierarchical}. 
The latter generate the entire graph at once \cite{simonovsky2018graphvae,shi2020graphaf,luo2021graphdf,martinkus2022spectre}.

Diffusion models have been adapted for molecular graph generation \cite{liu2023generative,qiu2023reconstructing,li2024regressor}. 
Some methods define a diffusion process directly on the discrete state space \cite{vignac2023digress,xu2024discrete}, which can be complex to accomplish. Others map discrete attributes to a continuous space and apply standard Gaussian diffusion \cite{huang2022graphgdp,lee2023exploring}.

BFNs have arisen recently to perform generation in continuous parameter distribution \cite{graves2023bayesian,song2024unified,qu2024molcraft}. The process starts with a simple prior distribution and iteratively refines its parameters via Bayesian inference to calculate the posterior from observed samples. 

\section{Preliminaries}

\subsection{Problem Definition}

A molecular graph $G$ is defined as $(X, A)$, where $X \in \{1, \cdots, K_{X}\}^{D}$ represents a matrix of $D$ atom features derived from $K_X$ classes, and $A \in \{1, \cdots, K_{A}\}^{D \times D}$ denotes an adjacency matrix encapsulating bond features from $K_A$ classes. Here $D$ is the maximum number of atoms in the considered molecules and let $N \leq D$ be the number of valid atoms.

Given a dataset $\mathcal{D} = \{G_i\}_{i=1}^{|\mathcal{D}|}$ sampled from an unknown data distribution $p_{\text{data}}(G)$, the goal of molecular graph generation is to learn a generative model $p_\theta(G)$ that can efficiently sample valid, diverse molecular graphs whose distribution matches $p_{\text{data}}$ in terms of structural and chemical statistics. For the discrete nature of molecular graphs, this task can be seen as a multi-class generative task, where we need to assign specific atomic or chemical bond types at each position subject to chemical validity constraints. In addition, the model may be guided towards desired properties $\mathbf{z}$ to achieve conditional molecular graph generation as a flexible extension.

\subsection{Bayesian Flow Networks}
\label{pre_BFN}

Bayesian Flow Networks can be conceptualized as a data transmission protocol between a sender and a receiver. Given data $\mathbf{x}=(x^{(1)}, \cdots, x^{(D)})$, let $\boldsymbol{\theta}=(\theta^{(1)}, \cdots, \theta^{(D)})$ be the parameters of a factorized \emph{input distribution}: $p_I(\mathbf{x}\,|\,\boldsymbol{\theta})=\prod_{d=1}^{D}p_I(x^{(d)}\,|\,\theta^{(d)})$. At each time $t \in [0,1]$, the sender corrupts $\mathbf{x}$ by a Gaussian channel $\alpha$ yielding a \emph{sender distribution}: $p_S(\mathbf{y}\,|\,\mathbf{x};\alpha)=\prod_{d=1}^{D}p_S(y^{(d)}\,|\,x^{(d)};\alpha)$. During the data transmission process, the input parameters $\boldsymbol{\theta}$ are input into a neural network $\Psi$, whose output is then parameterized as an \emph{output distribution}: $p_O(\mathbf{x}\,|\,\boldsymbol{\theta},t)=\prod_{d=1}^{D}p_O(x^{(d)}\,|\,\Psi(\boldsymbol{\theta},t))$. Since the receiver knows the form of the sender distribution but does not know $\mathbf{x}$, it integrates all possible $\mathbf{x}'$ and sender distributions, weighted by the output distribution, and obtains the \emph{receiver distribution}: $p_R(\mathbf{y}\,|\,\boldsymbol{\theta};t, \alpha)=\mathop{\mathbb{E}}\limits_{p_O(\mathbf{x}'|\boldsymbol{\theta};t)}p_S(\mathbf{y}\,|\,\mathbf{x}';\alpha)$. Subsequently, by applying Bayesian inference, \emph{Bayesian update function} $h(\cdot)$ is derived to repeatedly update the parameters: $\boldsymbol{\theta}' = h(\boldsymbol{\theta},\mathbf{y},\alpha)$ along a noise schedule, defined as a continuous-time Bayesian flow $p_F(\boldsymbol{\theta} \mid \mathbf{x};t)$. The \emph{Bayesian update distribution} is then defined by marginalizing out $\mathbf{y}$: $p_U(\boldsymbol{\theta}'\,|\,\boldsymbol{\theta},\mathbf{x};\alpha)=\mathop{\mathbb{E}}\limits_{p_S(\mathbf{y}|\mathbf{x};\alpha)}\delta(\boldsymbol{\theta}'-h(\boldsymbol{\theta},\mathbf{y},\alpha))$, where $\delta(\cdot)$ is the Dirac delta distribution. Now define the \emph{accuracy schedule} $\beta(t)$ as $\beta(t)=\int_{t'=0}^{t}\alpha(t')dt'$, and the \emph{Bayesian flow distribution} is the marginal distribution over input parameters at time $t$: $p_F(\boldsymbol{\theta}\,|\,\mathbf{x};t)=p_U(\boldsymbol{\theta}\,|\,\boldsymbol{\theta}_0;\mathbf{x};\beta(t))$.

\subsection{Optimal Transport and Quasi-Wasserstein Coupling on Graphs}
\label{sec:prelim_ot}

Optimal transport (OT) provides a geometry-aware way to compare two distributions by measuring the minimum cost required to move probability mass from one to the other~\cite{gabriel2019computational}. Consider a molecular graph $G=(X,A)$ with $N$ atoms, and let $C\in\mathbb{R}_+^{N\times N}$ be the geodesic ground-cost matrix induced by the shortest-path metric on $G$, i.e., $C_{ij}=\mathrm{dist}_{G}(i,j)$. Given two probability distributions $\mathbf{a},\mathbf{b}\in\Delta^{N-1}$ over atom indices, the entropically regularized OT cost is
$
\mathcal{W}_{\varepsilon}(\mathbf{a},\mathbf{b};C)
=
\min_{P\in\Pi(\mathbf{a},\mathbf{b})}
\langle P, C\rangle
+
\varepsilon\sum_{i,j} P_{ij}\big(\log P_{ij}-1\big)$, where $
\Pi(\mathbf{a},\mathbf{b})=\{P\ge 0\mid P\mathbf{1}=\mathbf{a},~P^\top\mathbf{1}=\mathbf{b}\}$. In practice, we solve the entropy-regularized OT and obtain the coupling $P$ via Sinkhorn iterations~\cite{cuturi2013sinkhorn}.

Quasi-Wasserstein (QW) extends this idea from comparing single scalar signals on atoms to comparing multi-class label fields on graphs. Specifically, for a categorical field with $K$ classes, we view the category probability map $P(\cdot,k)\in[0,1]^N$ as a mass distribution over atom indices for each class $k$, and compare prediction $P(\cdot,k)$ and target $Q(\cdot,k)$ by transporting mass within each class under the same geodesic cost $C$. The QW discrepancy then aggregates these per-class transport costs: 
$
\mathcal{D}_{\mathrm{QW}}(P,Q;C)
=
\frac{1}{K}\sum_{k=1}^{K}
\mathcal{W}_{\varepsilon}\!\big(\mathbf{a}_k,\mathbf{b}_k;C\big),
\mathbf{a}_k=\mathrm{Norm}\!\big(P(\cdot,k)\big),
\mathbf{b}_k=\mathrm{Norm}\!\big(Q(\cdot,k)\big),
$
where $\mathrm{Norm}(\cdot)$ denotes normalization to the simplex on valid atoms. In contrast to position-wise divergences, $\mathcal{D}_{\mathrm{QW}}$ penalizes errors according to how far probability mass must move along the graph, thereby encouraging local and topology-consistent corrections while discouraging long-range and structure-inconsistent reallocations.

\begin{figure*}
    \centering
    \includegraphics[width=1.\linewidth]{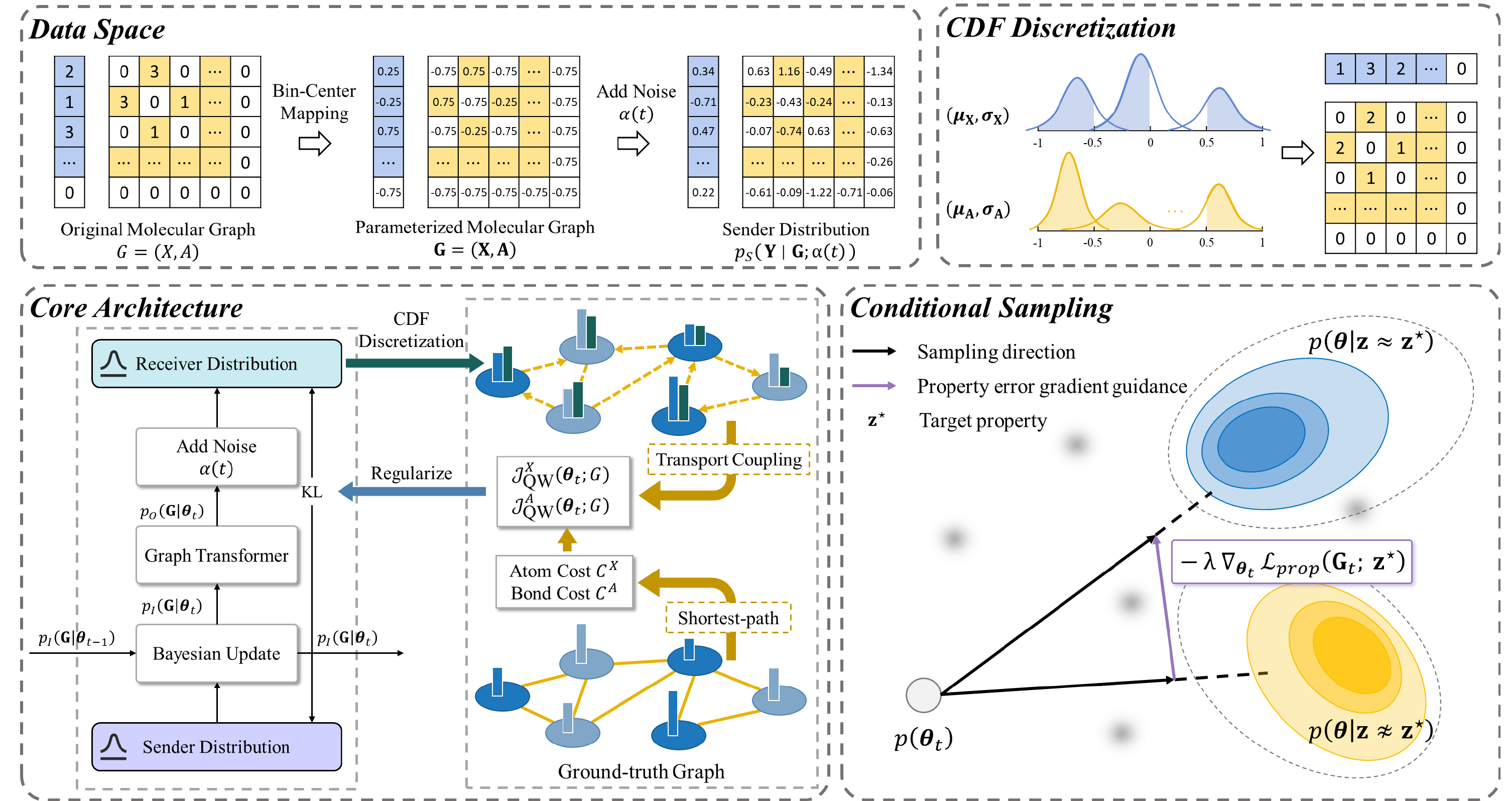}
    \caption{Overall framework of TopBF. \textbf{Data Space}: The procedure of parameterizing and adding noise to the molecular graph for the sender distribution. \textbf{CDF Discretization}: Obtain atom and bond categories from the output of $\Psi$ through CDF. \textbf{Core Architecture}: The structure of a BFN module at time $t$. And a QW transport coupling regularizes the training procedure. \textbf{Conditional Sampling}: PEG guidance for conditional generation.}
    \label{fig:model}
\end{figure*}

\section{Methodology} 

In this section, we describe how our MGG framework, \underline{\textbf{T}}ransp\underline{\textbf{o}}rt-Cou\underline{\textbf{p}}led \underline{\textbf{B}}ayesian \underline{\textbf{F}}lows (TopBF), generates molecular graphs via Bayesian flows over continuous distribution parameters, as illustrated in Figure~\ref{fig:model}. Pseudocodes for training and sampling are provided in Appendix~\ref{app:pseudocode}. 

\subsection{Parameterization of Molecular Graphs}

Since directly manipulating discrete data often leads to more complex operations to guarantee differentiability, a foundational step of our method involves mapping the discrete atom and bond features into continuous spaces. For a feature class $k \in \{1, \cdots, K\}$, we define a deterministic mapping to a continuous value $x_k \in [-1, 1]$:
\begin{equation}
    k_c = \frac{2k-1}{K} - 1, \quad k_l = k_c - \frac{1}{K},\quad k_r = k_c + \frac{1}{K},
\label{eq:class-mapping}
\end{equation}
where $k_l$, $k_c$ and $k_r$ represent the respective vectors for left boundaries, centers and right boundaries of the classes. Importantly, we assign the continuous value $x_k \coloneqq k_c$ to the $k$-th class. This mapping transforms the ground-truth graph $G$ into continuous representation $\mathbf{G}=(\mathbf{X}, \mathbf{A})$. This continuous representation serves as the target signal for the generative flow.

\subsection{Bayesian Flow Over Distribution Parameters}

\subsubsection{Input distribution in parameter space.} 
Instead of modeling $G$ directly as diffusion models, TopBF operate on parameters of a simple input distribution over continuous $\mathbf{G}$. For each of atoms and bonds we maintain a Gaussian input distribution:
\begin{align}
    p_I(\mathbf{X} \mid \boldsymbol{\theta}_\mathbf{X}) &= \mathcal{N}(\mathbf{X} \mid \boldsymbol{\mu}_\mathbf{X}, \rho_\mathbf{X}^{-1}\boldsymbol{I}), \\
    p_I(\mathbf{A} \mid \boldsymbol{\theta}_\mathbf{A}) &= \mathcal{N}(\mathbf{A} \mid \boldsymbol{\mu}_\mathbf{A}, \rho_\mathbf{A}^{-1}\boldsymbol{I}),
\end{align}
where $\boldsymbol{\theta} \overset{def}{=} (\boldsymbol{\mu}, \rho)$ consists of a mean vector $\mu \in \mathbb{R}^D$ and a scalar precision $\rho > 0$, and $\boldsymbol{I}$ is the $D \times D$ identity matrix. The process initiates with a standard normal prior $p_I(\cdot \mid \boldsymbol{\theta}_0) = \mathcal{N}(\cdot \mid \boldsymbol{0}, \boldsymbol{I})$ at time $t=0$. 
For brevity, we will use the integrated form in our formulas: $p_I(\mathbf{G} \mid \boldsymbol{\theta}) = \mathcal{N}(\mathbf{G} \mid \boldsymbol{\mu}, \rho^{-1}\boldsymbol{I})$, except in special circumstances. 

\subsubsection{Sender distribution: adding noise to the data.}
Given continuous $\mathbf{G}=(\mathbf{X}, \mathbf{A})$, the sender distribution plays a role forwarding noising data. For each time-dependent accuracy parameter $\alpha(t)$, the sender draws a noisy message:
\begin{equation}
    p_S(\mathbf{Y} \mid \mathbf{G}; \alpha(t)) = \mathcal{N}(\mathbf{Y} \mid \mathbf{G}, \alpha(t)^{-1}\boldsymbol{I}),
\label{eq:sender}
\end{equation}
where $\mathbf{Y} = (\mathrm{Y}^{(1)}, \cdots, \mathrm{Y}^{(D)})$ is a noisy observation of the true graph, whose noise level increases when $\alpha(t)$ is small.

\subsubsection{Bayesian update and Bayesian flow distribution.}
BFNs define a Bayesian update that combines the current input distribution $p_I(\mathbf{G} \mid \boldsymbol{\theta})$ with a noisy observation $\mathbf{Y}$ from the sender, yielding a posterior over $\mathbf{G}$ that can again be written as a Gaussian in closed form. Since both $p_I$ and $p_S$ are Gaussians with diagonal covariances, the Bayesian update for each of atoms and bonds reduces to \cite{murphy2007conjugate}:
\begin{equation} 
\label{eq:bayes-update}
\begin{aligned}
        \left\{
\begin{aligned}
    \rho_i &= \rho_{i-1} + \alpha, \\
    \boldsymbol{\mu}_i  &= \frac{\rho_{i-1} \boldsymbol{\mu}_{i-1} + \alpha \mathbf{Y}}{\rho_{i-1} + \alpha}.
\end{aligned}
\right.
\end{aligned}
\end{equation}
Repeating this update along a schedule of accuracies induces a stochastic trajectory $\{\boldsymbol{\theta}_t\}_{t \in [0,1]}$ in parameter space that converges towards the true graph $\mathbf{G}$ as $t \to 1$.

Rather than explicitly simulating all intermediate steps, there exists a \emph{Bayesian flow distribution} $p_F(\boldsymbol{\theta}\,|\,\mathbf{G}; t)$ describing the distribution of input parameters at time $t$ after aggregating all past Bayesian updates, as elaborated in Sec.~\ref{pre_BFN}.

Conceptually, the dynamics are defined on \emph{distribution parameters} $(\boldsymbol{\mu}, \rho)$ rather than on data itself. This is the key distinction of TopBF from standard diffusion models.

\subsection{Output and Receiver Distributions for Molecular Graphs}

The core neural network in TopBF takes as input the current parameter state $\boldsymbol{\theta}_t$ and time $t$, and outputs a prediction of the Gaussian noise used to generate the input mean. We denote the network by:
\begin{equation}
    \boldsymbol{\epsilon} = \Psi(\boldsymbol{\theta}_t, \mathbf{c}, t),
\end{equation}
where $\mathbf{c}$ represents any optional conditioning. The output $\boldsymbol{\epsilon}$ is split into two $D$-dimensional vectors $(\boldsymbol{\mu}_\epsilon, \ln \boldsymbol{\sigma}_\epsilon)$, which are then transformed into the mean and standard deviation $(\hat{\boldsymbol{\mu}}_\mathbf{G}, \hat{\boldsymbol{\sigma}}_\mathbf{G})$ of a Gaussian over the continuous data (formal derivation of the transformation in Appendix~\ref{app:eq7:derivation}):
\begin{equation}
    \begin{aligned}
        & \hat{\boldsymbol{\mu}}_\mathbf{G} = \left\{
    \begin{aligned}
        & \, \boldsymbol{0} & \text{if $t<t_{\text{min}}$},\\
        & \frac{\boldsymbol{\mu}}{\gamma(t)} - \sqrt{\frac{1-\gamma(t)}{\gamma(t)}}\boldsymbol{\mu}_\epsilon & \text{otherwise,}
    \end{aligned}
    \right. \\
    & \hat{\boldsymbol{\sigma}}_\mathbf{G} = \left\{
    \begin{aligned}
        & \, \boldsymbol{1} & \text{if $t<t_{\text{min}}$},\\
        & \sqrt{\frac{1-\gamma(t)}{\gamma(t)}}\text{exp}(\ln\boldsymbol{\sigma}_\epsilon) & \text{otherwise,}
    \end{aligned}
    \right.
    \end{aligned}
\label{eq:mu-sigma-x}
\end{equation}
where $\gamma(t) = \beta(t)/(1 + \beta(t))$ is determined by the accuracy schedule $\beta(t)$ of BFNs, and $t_{\min}$ is a small positive constant to avoid numerical issues at $t=0$.


\subsubsection{Truncated Gaussian CDF and per-class probabilities.}
While Eq.~\eqref{eq:mu-sigma-x} defines a continuous Gaussian distribution over $\mathbf{G}$, our ultimate goal is to make discrete predictions over $K$ classes. To bridge this gap, TopBF uses a truncated Gaussian cumulative distribution function (CDF) to compute the probability mass assigned to each discretization class.

For each dimension $d \in \{1,\cdots,D\}$ we define the Gaussian CDF:
\begin{equation}
    F(x \mid \mu_x^{(d)}, \sigma_x^{(d)}) = \frac{1}{2}\Bigg[ 1 + \text{erf} \left( \frac{x-\mu_x^{(d)}}{\sqrt{2}\sigma_x^{(d)}} \right) \Bigg],
\label{eq:gauss-cdf}
\end{equation}
where $\text{erf}(\cdot)$ is the error function defined as $\text{erf}(u)=\frac{2}{\sqrt{\pi}}\int_{0}^{u}e^{-t^2}dt$. Here truncate the support to the interval $[-1, 1]$ to align with the embedding range:
\begin{equation}
    \begin{aligned}
        & \mathcal{F}(x \,|\, \mu_x^{(d)}, \sigma_x^{(d)}) = \left\{
        \begin{aligned}
            & \, 0 & \text{if $ x \leq -1$ },\\
            & \, 1 & \text{if $ x \geq 1 $}, \\
            & F(x \,|\, \mu_x^{(d)}, \sigma_x^{(d)}) & \text{otherwise, }
        \end{aligned}
        \right.
    \end{aligned}
\end{equation}
Using the class boundaries $(k_l, k_r)$ from Eq.~\eqref{eq:class-mapping}, the probability of assigning dimension $d$ to category $k \in \{1, \cdots K\}^D$ is determined by the integral of the truncated Gaussian over the corresponding class:
\begin{equation}
    p_O^{(d)}(k \mid \boldsymbol{\theta}_t) = \mathcal{F}(k_r \mid \mu_x^{(d)}, \sigma_x^{(d)}) - \mathcal{F}(k_l \mid \mu_x^{(d)}, \sigma_x^{(d)}).
\label{eq:po-discrete}
\end{equation}
To achieve closed-form \emph{Bayesian update}, the output distribution over the $\mathbf{G}$ is independent across dimensions:
\begin{equation}
    p_O(\mathbf{G} \mid \boldsymbol{\theta}_t) = \prod_{d=1}^{D}p_O^{(d)}\Big( k(\mathrm{G}^{(d)}) \mid \boldsymbol{\theta}_t \Big),
\label{eq:po-factorised}
\end{equation}
where $k(x^{(d)})$ denotes the index of the class occupied by $x^{(d)}$. Similarly, we define the expected class-center representation under the current output distribution as:
\begin{equation}
    \begin{aligned}
        & \mathbb{E}[P(\boldsymbol{\theta}, t)] = \hat{\mathbf{k}}(\boldsymbol{\theta}, t) \\
        \overset{def}{=} & \left( \sum_{k=1}^{K}p^{(1)}(k \,|\, \boldsymbol{\theta}, t) k_{c}, \cdots, \sum_{k=1}^{K}p^{(D)}(k \,|\, \boldsymbol{\theta}, t) k_{c} \right).
    \end{aligned}
\end{equation}

\subsubsection{Receiver distribution as a mixture of Gaussians.}
The receiver distribution $p_R$ inverts the sender process in parameter space by describing the distribution over noisy messages $\mathbf{Y}$ implied by the current parameter state $\boldsymbol{\theta}_t$. Consequently, substituting Eqs.~\eqref{eq:sender} and~\eqref{eq:po-discrete} into the BFN formulation yields a mixture of Gaussians:
\begin{equation}
    \begin{aligned}
        & p_R(\mathbf{Y} \mid \boldsymbol{\theta}; t, \alpha(t)) = \mathop{\mathbb{E}}\limits_{p_O(\mathbf{G} \mid \boldsymbol{\theta}_t)} p_S(\mathbf{Y} \mid \mathbf{G}; \alpha(t)), \\
        & = \prod_{d=1}^D \sum_{k=1}^{K}p_O^{(d)}(k \mid \boldsymbol{\theta}_t) \mathcal{N}(\mathrm{Y}^{(d)} \mid k_c, \alpha(t)^{-1}).
    \end{aligned}
\label{eq:receiver-discrete}
\end{equation}
The receiver plays the role of the \emph{reverse} channel in the flow. During training, we encourage the receiver distribution to match the sender distribution applied to $\mathbf{G}$, while during sampling it serves to generate pseudo-observations that drive the Bayesian updates in Eq.~\eqref{eq:bayes-update}. Notably, TopBF applies the above construction independently to atoms and bonds by using $K = K_X$ and $K = K_A$, respectively.

\subsection{Topology-aware Optimal Transport Matching}
\label{sec:topo_ot}

Although the position-wise factorization in Eq.~\eqref{eq:po-factorised} enables closed-form Bayesian updates, it also yields largely local and index-wise supervision that does not explicitly encode graph topology. However, categorical errors are not equally severe, since swapping mass within a local neighborhood is structurally less disruptive than relocating it across distant substructures \cite{frogner2015learning,cheng2024quasi}. To bridge this gap, we introduce an OT coupling that regularizes predictions under molecule's intrinsic geometry.

\subsubsection{Graph-induced Ground Cost.}
We define the atom-level ground cost by the shortest-path metric: 
\begin{equation}
    C^{X}_{ij} \;=\; \mathrm{dist}_{G}(i,j),\qquad i,j\in\{1,\cdots,N\}.
\label{eq:cost_X}
\end{equation}
For bonds, we define an bond-level cost $C^{A}$ using an endpoint-based proxy consistent with our implementation, e.g., for two undirected bonds $e=(i,j)$ and $e'=(u,v)$,
\begin{equation}
\begin{aligned}
C^{A}_{e,e'} = \min \big\{ & \mathrm{dist}_{G}(i,u), \mathrm{dist}_{G}(i,v), \\
& \mathrm{dist}_{G}(j,u), \mathrm{dist}_{G}(j,v) \big\}.
\label{eq:cost_A}
\end{aligned}
\end{equation}
\begin{theorem}[Endpoint proxy matches line-graph geodesics]
\label{thm:line_graph}
Let $\mathcal{L}(G)$ be the line graph of $G$ whose vertices correspond to bonds and edges connect bonds that share an endpoint. For any two distinct undirected bonds $e,e'$,
\[
\mathrm{dist}_{\mathcal{L}(G)}(e,e') \;=\; 1 + C^{A}_{e,e'} .
\]
\end{theorem}
Thus, $C^{A}$ differs from the exact line-graph geodesic only by an additive constant and induces the same optimal OT coupling (proof in Appendix~\ref{app:qw:edge_cost}).

\subsubsection{Optimal Transport over Categorical Mass.}
We make topology awareness a first-class objective by minimizing the transport work needed to align predicted categorical mass with the ground truth~\cite{peyre2016gromov}.
At time $t$, a sample $\boldsymbol{\theta}_t\sim p_F(\boldsymbol{\theta} \,|\, \mathbf{G};t)$ yields categorical probabilities from Eq.~\eqref{eq:po-factorised}, with targets $Q_X,Q_A$. We form class-wise mass fields by normalizing $P_X(\cdot,k)$ and $P_A(\cdot,k)$ on valid indices and only aggregate over classes that appear in the true graph $G$: $\mathcal{K}_X^{+}(G)=\{k \mid \exists\, i \ \text{s.t.}\ X_i=k\}$ and $\mathcal{K}_A^{+}(G)=\{k \mid \exists\, e \ \text{s.t.}\ A_e=k\}$. Thus we define the QW transport costs
\begin{align}
\mathcal{J}^{X}_{\mathrm{QW}}(\boldsymbol{\theta}_t;G)
&=
\frac{1}{|\mathcal{K}_X^{+}(G)|}
\sum_{k\in\mathcal{K}_X^{+}(G)}
\!\mathcal{W}_{\varepsilon}\!\big(\mathbf{a}_k^X,\mathbf{b}_k^X;C^{X}\big),
\label{eq:qw_X}
\\
\mathcal{J}^{A}_{\mathrm{QW}}(\boldsymbol{\theta}_t;G)
&=
\frac{1}{|\mathcal{K}_A^{+}(G)|}
\sum_{k\in\mathcal{K}_A^{+}(G)}
\!\mathcal{W}_{\varepsilon}\!\big(\mathbf{a}_k^A,\mathbf{b}_k^A;C^{A}\big),
\label{eq:qw_A}
\end{align}
where $\mathcal{W}_{\varepsilon}$ is the entropic OT cost in Sec.~\ref{sec:prelim_ot}; for bonds we exclude the  invalid bond class from $\mathcal{K}_A^{+}(G)$ to avoid trivial domination. The overall training objective then minimizes the expected transport work induced by the receiver predictions, 

\begin{theorem}[Hard-label limit of class-wise transport]
\label{thm:em}
Fix a class $k$ and suppose $\mathbf{a}_k=\frac{1}{r}\sum_{\ell=1}^{r}\mathbf{e}_{i_\ell}$ and $\mathbf{b}_k=\frac{1}{r}\sum_{\ell=1}^{r}\mathbf{e}_{j_\ell}$ are uniform sums of $r$ Diracs. Then
\[
\mathcal{W}_{0}(\mathbf{a}_k,\mathbf{b}_k;C)
=
\frac{1}{r}\min_{\pi\in\mathfrak{S}_r}\sum_{\ell=1}^{r} C_{i_\ell,j_{\pi(\ell)}}.
\]
\end{theorem}
Proof is in Appendix~\ref{proof:theorem_2}.

\begin{table*}
  \centering
  \renewcommand \arraystretch{1.}
  \begin{tabular}{cccccccccc}
    \toprule

    \multirow{2}{*}{Method} & \multicolumn{4}{c}{QM9$_{\,\, (|N|\leq\text{9})}$} & \multicolumn{4}{c}{ZINC250k$_{\,\, (|N|\leq\text{38})}$} & \multirow{2}{*}{\makecell[c]{Sampling \\ Steps}} \\
    \cmidrule(lr){2-5} \cmidrule(lr){6-9}
     & $\text{Valid(\%)}\uparrow$ & Unique$\uparrow$ & FCD$\downarrow$ & NSPDK$\downarrow$ & $\text{Valid(\%)}\uparrow$ & Unique$\uparrow$ & FCD$\downarrow$ & NSPDK$\downarrow$ \\
    \midrule
    EDP-GNN & 47.52 & \underline{99.25} & 2.680 & 0.0046 & 82.97 & 99.79 & 16.737 & 0.0485 & 1000 \\
    GDSS & 95.72 & 98.46 & 2.900 & 0.0033 & 97.01 & 99.64 & 14.656 & 0.0195 & 1000 \\
    DiGress & 98.19 & 96.20 & \underline{0.095} & 0.0003 & 94.99 & 99.97 & 3.482 & 0.0021 & 500 \\
    LGD-large & 99.13 & 96.82 & 0.104 & \underline{0.0002} & - & - & - & - & 1000 \\
    GruM & \underline{99.69} & 96.90 & 0.108 & \underline{0.0002} & 98.65 & 99.97 & 2.257 & 0.0015 & 1000 \\
    DeFoG & 99.30 & 96.30 & 0.120 & - & 99.22 & \underline{99.99} & \underline{1.425} & \textbf{0.0008} & 1000 \\
    SID & 99.67 & 95.66 & 0.504 & \textbf{0.0001} & \textbf{99.50} & 99.84 & 2.010 & 0.0021 & 500 \\
    \midrule
    $\text{TopBF}$ & \textbf{99.74} & \textbf{99.27} & \textbf{0.093} & \underline{0.0002} & \underline{99.37} & \textbf{100.00} & \textbf{1.392} & \textbf{0.0008} & \textbf{200} \\
    
    \bottomrule
  \end{tabular}
  \caption{Performance comparison on QM9 and ZINC250k datasets. The best results are highlighted in \textbf{bold} and the second best are \underline{underlined}. Null values (-) in criteria indicate that statistics are unavailable from the original paper.}
  \label{tab:1}
\end{table*}

\subsection{Training Objective}
\label{BFN Loss}

The BFN loss can be interpreted either as the number of nats required to transmit the data through the Bayesian flow, or equivalently as the negative variational lower bound over message sequences. Concretely, we sample $t\sim\mathcal{U}(0,1)$ and $\boldsymbol{\theta}_t\sim p_F(\boldsymbol{\theta}\mid \mathbf{G};t)$, and minimize the expected divergence between the sender distribution applied to the ground-truth discretized graph and the receiver distribution implied by $\boldsymbol{\theta}_t$: 
\begin{equation}
\!\mathcal{L}_{\mathbf{G}}
\!=\!
\mathop{\mathbb{E}}\limits_{t,\boldsymbol{\theta}_t}
\Big[
{\mathrm{KL}}\!\Big(
p_S(\mathbf{Y} | \mathbf{G};\alpha(t))
\big\Vert
p_R(\mathbf{Y} | \boldsymbol{\theta}_t; t,\alpha(t))
\Big)
\Big].
\label{eq:bfn_cont_time_kl}
\end{equation}

In TopBF, $p_S$ and $p_R$ are Gaussians with diagonal covariance and we adopt the accuracy schedule $\alpha(t)=\sigma_1^{-2t}$. For a single continuous feature, it makes the KL divergence proportional to a squared error between the true value and the receiver mean. For discretized atom and bond features, we represent each category by $k_c$ in the one-dimensional continuous space and use the expectation of these centers under $p_O$. Here it yields the continuous-time losses for atoms and bonds (Appendix~\ref{app:loss} for a detailed derivation):
\begin{equation}
\label{eq:atom-loss}
    \!\!\mathcal{L}_{\mathbf{X}}
    = - \ln \sigma_{1} \,
    \mathop{\mathbb{E}}\limits_{t \sim \mathcal{U}(0,1),\, p_F(\boldsymbol{\theta} \mid \mathbf{X}; t)}
    \!\!\left[
        \frac{\big\| \mathbf{X} - \hat{\mathbf{k}}_{\mathbf{X}} (\boldsymbol{\theta}, t) \big\|_2^2}
             {\sigma_{1}^{2t}}
    \right],
\end{equation}
where $\sigma_{1}$ is the target standard deviation of the input distribution at $t=1$. An analogous loss $\mathcal{L}_A$ is defined for bond features and the base objective is $ \mathcal{L}_{\mathbf{G}} = \mathcal{L}_{\mathbf{X}} + \mathcal{L}_{\mathbf{A}}$.

To inject explicit topology awareness into training, we further minimize the expected Quasi-Wasserstein transport work defined in Sec.~\ref{sec:topo_ot}: 
\begin{equation}
\mathop{\mathbb{E}}\limits_{t\sim\mathcal{U}(0,1),\,\boldsymbol{\theta}_t\sim p_F(\boldsymbol{\theta}\mid \mathbf{G};t)}
\Big[
\mathcal{J}^{X}_{\mathrm{QW}}(\boldsymbol{\theta}_t; \mathbf{G})
+
\mathcal{J}^{A}_{\mathrm{QW}}(\boldsymbol{\theta}_t; \mathbf{G})
\Big].
\label{eq:total_topo_objective}
\end{equation}
This joint objective preserves the original Bayesian flow formulation while encouraging predictions that align with the ground-truth graph through low-cost geodesic transport.

\begin{table}
  \centering
  \renewcommand \arraystretch{1.}
  \begin{tabular}{ccccc}
    \toprule

    \multicolumn{2}{c}{Target} & $\mu\downarrow$ & HOMO$\downarrow$ & $\mu$ \& HOMO$\downarrow$ \\
    \midrule
    \multicolumn{2}{c}{Unconditional} & $\text{1.69}_{\pm\text{0.05}}$ & $\text{0.96}_{\pm\text{0.01}}$ & $\text{1.73}_{\pm\text{0.02}}$ \\
    \midrule
    \multirow{3}{*}{+PEG} & $\lambda=0.5$ & $\text{0.92}_{\pm\text{0.04}}$ & $\text{0.71}_{\pm\text{0.02}}$ & $\text{0.93}_{\pm\text{0.03}}$ \\
     & $\lambda=1.0$ & $\textbf{0.63}_{\pm\text{0.04}}$ & $\textbf{0.51}_{\pm\text{0.01}}$ & $\textbf{0.77}_{\pm\text{0.03}}$ \\
     & $\lambda=1.5$ & $\text{1.55}_{\pm\text{0.07}}$ & $\text{0.83}_{\pm\text{0.03}}$ & $\text{1.29}_{\pm\text{0.06}}$ \\
    \bottomrule
  \end{tabular}
  \caption{Comparison on mean absolute error (MAE) between unconditional generation and PEG guidance generation.}
  \label{tab:qm9-conditional}
\end{table}

\subsection{Conditional Generation}

TopBF supports conditional generation without retraining the unconditional flow model. Integrating property control via classifier-free \cite{ho2021classifier} or regressor-free \cite{li2024regressor} guidance often necessitates retraining models from scratch for each target property, which limits their flexibility across different tasks. Instead, we train a separate regressor and apply property-error gradient (PEG) guidance during sampling, which remains flexible for changing targets or composing multiple objectives.

Let $\mathcal{D}_{\text{prop}}=\{(G_i,\mathbf{z}_i)\}_{i=1}^{M}$ be a property-labeled dataset with property $\mathbf{z}\in\mathbb{R}^{d_z}$. We train a lightweight regressor $R_{\varphi}$ on noisy graphs produced by the same sender corruption used in TopBF so that the regressor predicts the clean property from noisy inputs:
\begin{equation}
    \hat{\mathbf{z}}_t = R_\varphi(\mathbf{X}_t, \mathbf{A}_t, t),
\end{equation}
\begin{equation}
    \mathcal{L}_{\text{reg}}
    =
    \mathop{\mathbb{E}}\limits_{(G, \mathbf{z}) \sim \mathcal{D}_{\text{prop}}}
    \,
    \mathop{\mathbb{E}}\limits_{t \sim \mathcal{U}(0,1)}
    \,
    \mathop{\mathbb{E}}\limits_{\mathbf{Y} \sim p_S(\cdot \mid \mathbf{G}; \alpha(t))}
    \big[
        \| \hat{\mathbf{z}}_t - \mathbf{z} \|_2^2
    \big].
\end{equation}
Consequently, $R_\varphi$ learns to estimate properties consistently along the BFN corruption path.

During sampling, we aim to obtain molecules with property close to specified target $\mathbf{z}^\star$. Given the current expected class centers under $p_O(\cdot \mid \boldsymbol{\theta}_t)$, we compute $\hat{\mathbf{z}}_t = R_\varphi(\mathbf{G}_t, t)$ and define a property error: 
\begin{equation}
    \mathcal{L}_{\text{prop}}(\mathbf{G}_t; \mathbf{z}^\star)
    =
    \| \hat{\mathbf{z}}_t - \mathbf{z}^\star \|_2^2.
\end{equation}
Before applying the Bayesian update, we take a small gradient step in parameter space
\begin{equation}
    \tilde{\boldsymbol{\theta}}_t
    =
    \boldsymbol{\theta}_t - \lambda \nabla_{\boldsymbol{\theta}_t} \mathcal{L}_{\text{prop}}(\mathbf{G}_t; \mathbf{z}^\star),
\end{equation}
where $\lambda \geq 0$ controls the guidance strength and $\lambda = 0$ recovers the unconditional sampling.

\begin{table*}
  \centering
  \renewcommand \arraystretch{1.}
  \begin{tabular}{cccccccccc}
    \toprule

    \multicolumn{2}{c}{QW on} & \multicolumn{4}{c}{QM9$_{\,\, (|N|\leq\text{9})}$} & \multicolumn{4}{c}{ZINC250k$_{\,\, (|N|\leq\text{38})}$} \\
    \cmidrule(lr){1-2} \cmidrule(lr){3-6} \cmidrule(lr){7-10}
     Atoms & Bonds & $\text{Valid(\%)}\uparrow$ & Unique$\uparrow$ & FCD$\downarrow$ & NSPDK$\downarrow$ & $\text{Valid(\%)}\uparrow$ & Unique$\uparrow$ & FCD$\downarrow$ & NSPDK$\downarrow$ \\
    \midrule
     \ding{56} & \ding{56} & 98.63 & 97.85 & 0.143 & 0.0003 & 94.75 & 99.03 & 3.959 & 0.0031 \\
     \ding{52} & \ding{56} & 98.79 & 99.13 & \underline{0.102} & 0.0003 & 95.61 & \textbf{100.00} & \textbf{1.347} & 0.0011 \\
     \ding{56} & \ding{52} & \textbf{99.81} & \underline{99.05} & 0.107 & \textbf{0.0002} & \underline{99.33} & 99.99 & 2.016 & \underline{0.0009} \\
     \ding{52} & \ding{52} & \underline{99.74} & \textbf{99.27} & \textbf{0.093} & \textbf{0.0002} & \textbf{99.37} & \textbf{100.00} & \underline{1.392} & \textbf{0.0008} \\
    
    \bottomrule
  \end{tabular}
  \caption{Ablation study on Quasi-Wasserstein transport coupling. The best results are highlighted in \textbf{bold} and the second best are \underline{underlined}.}
  \label{tab:ablation_qw}
\end{table*}

\begin{figure}
    \centering
    \includegraphics[width=1.\linewidth]{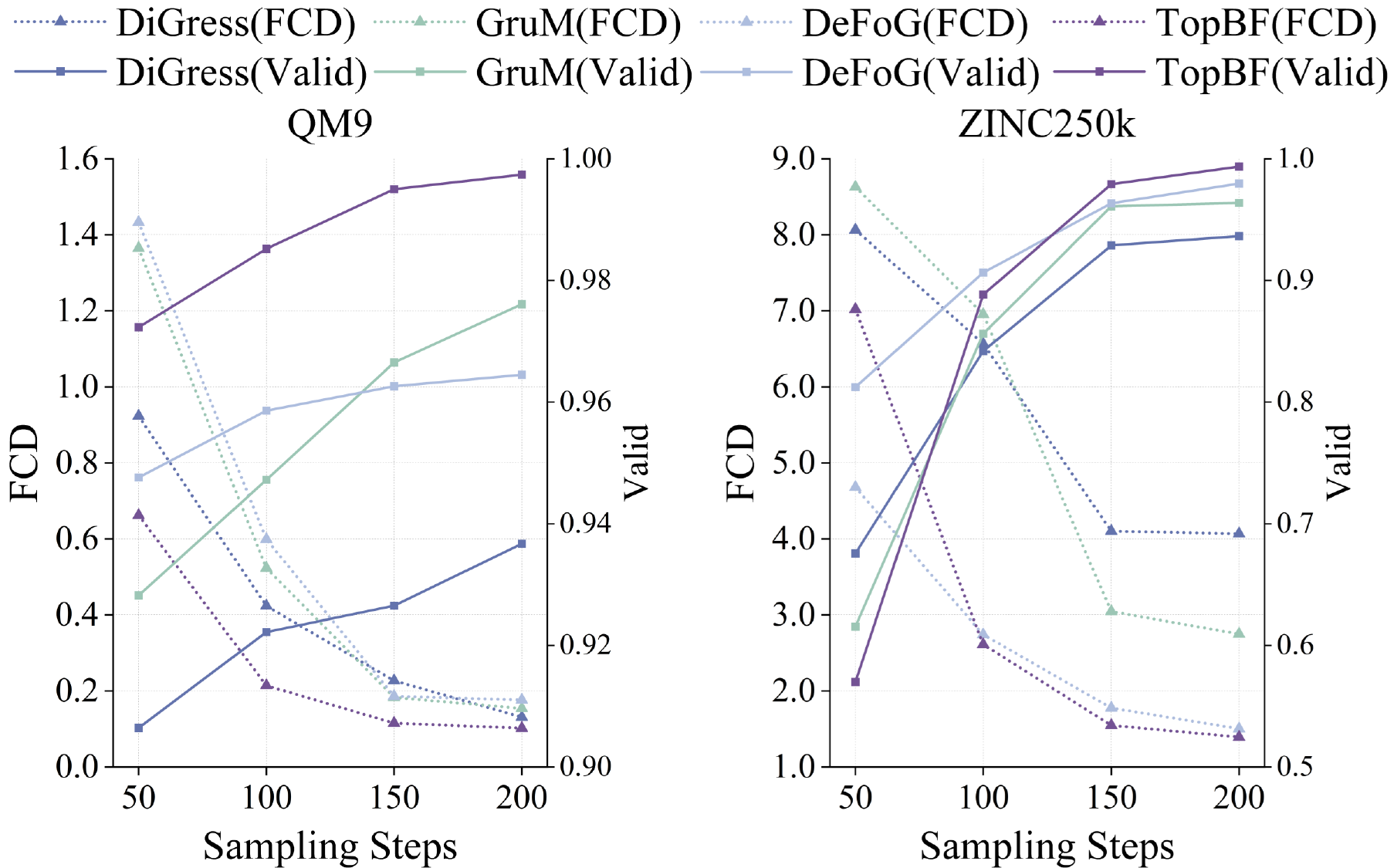}
    \caption{Sampling efficiency comparison between TopBF and baselines on QM9 and ZINC250k datasets.}
    \label{fig:sampling}
\end{figure}

\section{Experiment}

We experimentally validate the performance of our method in molecular graph generation task as well as sampling efficiency comparison against representative baselines.

\subsection{Experimental Setup} 
\subsubsection{Datasets and Evaluation Metrics} 
We choose QM9 \cite{ramakrishnan2014quantum} and ZINC250k \cite{irwin2012zinc} for molecular graph generation, with the evaluation setting following \cite{jo2022score}: Validity without correction and uniqueness on 10,000 generated molecules. Fréchet ChemNet Distance (FCD) \cite{preuer2018frechet} measures the distance between training and generated sets using the activations from the penultimate layer of the ChemNet. Additionally, Neighborhood Subgraph Pairwise Distance Kernel (NSPDK) maximum mean discrepancy (MMD) \cite{costa2010fast} evaluates the MMD between the generated molecules and test molecules which takes into account both the atom and bond features for evaluation.

\subsubsection{Baselines}
We compare our model against several excellent baselines in MGG. 
EDP-GNN \cite{niu2020permutation} models graphs with a permutation invariant approach. 
GDSS \cite{jo2022score} is a score-based diffusion model. 
DiGress \cite{vignac2023digress} generates graphs using a discrete diffusion process. 
LGD \cite{zhou2024unifying} provides a unified framework for all-level graph generation and prediction via latent diffusion.
GruM \cite{jo2024graph} models the generative process as a mixture of endpoint-conditioned diffusion.
DeFoG \cite{qinmadeira2024defog} using a discrete flow matching framework that defines continuous-time probability paths over categorical attributes.
SID \cite{boget2025simple} assumes conditional independence between intermediate states to simplify discrete diffusion.

\subsection{Molecular Graph Generation}
In our experiments, TopBF demonstrates superior performance for most metrics on both QM9 and ZINC250k, as shown in Table~\ref{tab:1}, outperforming previous flow-based, flow matching, and diffusion models. Specifically, overall leading on FCD and NSPDK indicate the molecules generated by TopBF closely align with the distributions of the relevant molecular datasets. Furthermore, the highest score on uniqueness manifests the best diversity performance and exploration capability during generation, which further highlights the deficiency of the methods treating training objective as a regression task. Since all methods attain 100\% validity after molecular correction, recent methods including TopBF meet industrial requirements. Additionally, Table~\ref{tab:1} reveals that TopBF achieves superior effects with significantly fewer sampling steps. We also present randomly selected molecular samples in Appendix~\ref{app:vis} for visualization. 

\subsection{Conditional Generation}

We assess the controllability of TopBF on QM9 by conditioning on graph-level properties, following the setting of DiGress \cite{vignac2023digress}. From the QM9 test split we randomly select $100$ molecules, record their dipole moment $\mu$ and HOMO energy, and use each pair $(\mu^\star,\mathrm{HOMO}^\star)$ as a target. For every target we generate $10$ molecules unconditionally and conditionally with PEG guidance, respectively. $\mu$ and HOMO are computed with Psi4 \cite{smith2020psi4} from conformers built with RDKit \footnote{\url{https://www.rdkit.org}}, and mean absolute error (MAE) is reported between the targets and the estimated properties for above three tasks. We also carry out sensitivity analysis on $\lambda$. As shown in Table~\ref{tab:qm9-conditional}, TopBF with PEG guidance consistently yields lower MAE than the unconditional sampler, demonstrating effective property control. PEG with $\lambda=1.0$ performs best, as we infer that $\lambda=0.5$ guides little during sampling and guidance strength with $\lambda=1.5$ invades the original gradient to generate valid molecules, affecting the property values conversely.

\subsection{Sampling Efficiency and Effect}
We conduct a sampling efficiency experiment to compare the sampling velocity against representative baselines. As manifested in Figure~\ref{fig:sampling}, our method exhibits exceptional performance during initial 100 steps. Scores could be further improved if sampling with steps longer than 200. However, we choose 200 steps as the best accuracy-efficiency trade-off. This characteristic renders TopBF well-suited for generating high-quality molecular samples while maintaining competitive in inference time.

\subsection{Ablation on Quasi-Wasserstein Coupling}
\label{sec:ablation_qw}

We further conduct an ablation study to isolate the effect of the proposed QW transport coupling on atom types and bond types. All variants share the same architecture, training schedule, and sampling steps, differing only in whether the topology-aware transport objective is activated on atoms and/or bonds. As reported in Table~\ref{tab:ablation_qw}, applying QW to both atoms and bonds achieves the best overall performance. Moreover, the two partial variants reveal complementary behaviors, where enabling QW on atoms mainly enhances distributional fidelity reflected by FCD, while enabling QW on bonds primarily strengthens validity and graph-structural alignment reflected by NSPDK. Removing QW leads to a uniform degradation, indicating that topology-aware transport coupling provides an effective structural learning signal beyond CDF-aligned Bayesian flow objective with the default $L_2$ regularization, and that jointly coupling atoms and bonds yields the most reliable gains.

\section{Conclusion}
\label{sec:conclusion}

In this paper, we introduce \textbf{TopBF}, a transport-coupled Bayesian Flow framework for molecular graph generation. TopBF operates in continuous parameter distributions and employs a CDF-based output channel to directly compute category probabilities, aligning the training objective with the discrete discretization used during sampling procedure, rather than fitting arbitrary continuous embeddings. To further inject topology awareness without breaking closed-form Bayesian updates, we formulate a Quasi-Wasserstein transport coupling under geodesic costs, encouraging topology-consistent alignment of atom and bond categorical mass on molecular graphs. Empirically, TopBF achieves state-of-the-art results on standard molecular graph generation benchmarks with efficient sampling of the minimum steps.

\bibliographystyle{named}
\bibliography{ref}

\appendix
\newpage




\clearpage
\onecolumn

\section{Derivation of Eq.~(7)}
\label{app:eq7:derivation}

We derive how the network outputs $(\boldsymbol{\mu}_\epsilon, \ln\boldsymbol{\sigma}_\epsilon)$ are mapped to the mean and standard deviation $(\boldsymbol{\mu}_\mathbf{G}, \boldsymbol{\sigma}_\mathbf{G})$ of the Gaussian over $\mathbf{G}$ used in Eq.~(7).

Consider one scalar dimension $\mathrm{G}^{(d)}$ of the parameterized continuous graph representation $\mathbf{G}$, and let $\mu_t^{(d)}$ be the corresponding entry of the input mean at time $t$.
From the Bayesian flow over parameters, the marginal distribution of $\mu_t^{(d)}$ conditioned on $\mathrm{G}^{(d)}$ is
\begin{equation}
    \mu_t^{(d)} \mid \mathrm{G}^{(d)}, t
    \sim
    \mathcal{N}\!\big(
        \gamma(t)\,\mathrm{G}^{(d)},
        \gamma(t)\bigl(1-\gamma(t)\bigr)
    \big),
\end{equation}
where $\gamma(t)$ is the function defined in the main text.

Therefore there exists $\epsilon^{(d)} \sim \mathcal{N}(0,1)$ such that
\begin{equation}
    \mu_t^{(d)}
    =
    \gamma(t)\,\mathrm{G}^{(d)}
    +
    \sqrt{\gamma(t)\bigl(1-\gamma(t)\bigr)}\,
    \epsilon^{(d)}.
\end{equation}
Solving for $\mathrm{G}^{(d)}$ gives
\begin{equation}
    \mathrm{G}^{(d)}
    =
    \frac{\mu_t^{(d)}}{\gamma(t)}
    -
    \sqrt{\frac{1-\gamma(t)}{\gamma(t)}}\,
    \epsilon^{(d)}.
    \label{eq:app-G-from-eps}
\end{equation}

The core network at time $t$ takes $\boldsymbol{\theta}_t$ and optional conditioning as input and outputs
\begin{equation}
    \boldsymbol{\epsilon}
    =
    \Psi(\boldsymbol{\theta}_t, \mathbf{c}, t),
\end{equation}
which is split into two $D$-dimensional vectors $(\boldsymbol{\mu}_\epsilon, \ln\boldsymbol{\sigma}_\epsilon)$.
For each dimension $d$ we model the noise as
\begin{equation}
    \begin{aligned}
        \left\{
    \begin{aligned}
        & \epsilon^{(d)} \mid \boldsymbol{\theta}_t, t \sim \mathcal{N}\!\big( \mu_\epsilon^{(d)}, (\sigma_\epsilon^{(d)})^2 \big), \\
        & \sigma_\epsilon^{(d)} = \exp\bigl(\ln\sigma_\epsilon^{(d)}\bigr), \\
    \end{aligned}
    \right.
    \end{aligned}
\end{equation}

Substituting this into Eq.~\eqref{eq:app-G-from-eps}, we obtain an affine transform of a Gaussian random variable:
\begin{equation}
    \mathrm{G}^{(d)} \mid \boldsymbol{\theta}_t, t
    =
    \frac{\mu_t^{(d)}}{\gamma(t)}
    -
    \sqrt{\frac{1-\gamma(t)}{\gamma(t)}}\,
    \epsilon^{(d)}.
\end{equation}
If $Z \sim \mathcal{N}(\mu_Z, \sigma_Z^2)$ and $a,b \in \mathbb{R}$, then $a + bZ \sim \mathcal{N}(a + b\mu_Z, b^2\sigma_Z^2)$.  
Applying this with
\[
    Z = \epsilon^{(d)},\quad
    a = \frac{\mu_t^{(d)}}{\gamma(t)},\quad
    b = -\sqrt{\tfrac{1-\gamma(t)}{\gamma(t)}},
\]
we obtain
\begin{equation}
    \begin{aligned}
        \left\{
    \begin{aligned}
        & \mathrm{G}^{(d)} \mid \boldsymbol{\theta}_t, t \sim \mathcal{N}\!\Big( \mu_\mathrm{G}^{(d)}, (\sigma_\mathrm{G}^{(d)})^2 \Big), \\
        & \mu_\mathrm{G}^{(d)} = \frac{\mu_t^{(d)}}{\gamma(t)} - \sqrt{\frac{1-\gamma(t)}{\gamma(t)}}\,\mu_\epsilon^{(d)}, \\
        & (\sigma_\mathrm{G}^{(d)})^2 = \frac{1-\gamma(t)}{\gamma(t)}\, (\sigma_\epsilon^{(d)})^2,
    \end{aligned}
    \right.
    \end{aligned}
\end{equation}
and hence
\begin{equation}
    \sigma_\mathrm{G}^{(d)}
    =
    \sqrt{\frac{1-\gamma(t)}{\gamma(t)}}\,
    \sigma_\epsilon^{(d)}
    =
    \sqrt{\frac{1-\gamma(t)}{\gamma(t)}}\,
    \exp\bigl(\ln\sigma_\epsilon^{(d)}\bigr).
\end{equation}

Stacking over all dimensions $d=1,\dots,D$ and fixing $\boldsymbol{\mu}_\mathbf{G} = \mathbf{0}, \boldsymbol{\sigma}_\mathbf{G} = \mathbf{1}$ to avoid numerical issues when $\gamma(t)$ is extremely small, we yields:
\begin{equation}
    \begin{aligned}
        & \boldsymbol{\mu}_\mathbf{G} = \left\{
    \begin{aligned}
        & \, \boldsymbol{0} & \text{if $t<t_{\text{min}}$},\\
        & \frac{\boldsymbol{\mu}}{\gamma(t)} - \sqrt{\frac{1-\gamma(t)}{\gamma(t)}}\boldsymbol{\mu}_\epsilon & \text{otherwise,}
    \end{aligned}
    \right. \\
    & \boldsymbol{\sigma}_\mathbf{G} = \left\{
    \begin{aligned}
        & \, \boldsymbol{1} & \text{if $t<t_{\text{min}}$},\\
        & \sqrt{\frac{1-\gamma(t)}{\gamma(t)}}\text{exp}(\ln\boldsymbol{\sigma}_\epsilon) & \text{otherwise.}
    \end{aligned}
    \right.
    \end{aligned}
\end{equation}

\section{From KL flow loss to squared-error objective}
\label{app:loss}

Let $\mathbf{X}\in\mathbb{R}^{D}$ be the continuous parameterization of the atom features of a graph $G$
(the bond case is analogous), and let $\mathbf{Y}\in\mathbb{R}^{D}$ denote the noisy message received
from the sender. The continuous-time flow objective used by TopBF on $\mathbf{X}$ is
\begin{equation}
\mathcal{L}_{\text{flow}}(\mathbf{X})
=
\int_0^1 
\mathop{\mathbb{E}}\limits_{\boldsymbol{\theta}_t \sim p_F(\boldsymbol{\theta}\mid \mathbf{X};t)} \Big[ D_{\mathrm{KL}}\big( p_S(\mathbf{Y} \mid \mathbf{X};\alpha(t))
\Vert
p_R(\mathbf{Y} \mid \boldsymbol{\theta}_t; t,\alpha(t)) \big) \Big] \,\mathrm{d}t.
\label{eq:cts-flow-kl}
\end{equation}

By construction (Sec.~3.2), at each time $t$ the sender and receiver for atoms are Gaussians with the
same covariance:
\begin{align}
p_S(\mathbf{Y} \mid \mathbf{X};\alpha(t))
&=
\mathcal{N}\big(\mathbf{Y} \mid \mathbf{X},\,\alpha(t)^{-1}\mathbf{I}\big), \\
p_R(\mathbf{Y} \mid \boldsymbol{\theta}_t; t,\alpha(t))
&=
\mathcal{N}\big(\mathbf{Y} \mid \hat{\mathbf{k}}_{\mathbf{X}}(\boldsymbol{\theta}_t,t),\,\alpha(t)^{-1}\mathbf{I}\big),
\end{align}
where $\hat{\mathbf{k}}_{\mathbf{X}}(\boldsymbol{\theta}_t,t)$ is the vector of expected bin centres of
the output distribution $p_O$ at time $t$.

For two Gaussians with identical covariance $\Sigma = \alpha^{-1}\mathbf{I}$,
\begin{equation}
\begin{aligned}
& D_{\mathrm{KL}}\big(
\mathcal{N}(\mathbf{m}_1,\Sigma)\ \Vert\ \mathcal{N}(\mathbf{m}_2,\Sigma)
\big)\\
& =
\frac{1}{2}(\mathbf{m}_1-\mathbf{m}_2)^\top \Sigma^{-1} (\mathbf{m}_1-\mathbf{m}_2) \\
& =
\frac{\alpha}{2}\,\|\mathbf{m}_1-\mathbf{m}_2\|_2^2.
\end{aligned}
\end{equation}
Setting $\mathbf{m}_1=\mathbf{X}$ and $\mathbf{m}_2=\hat{\mathbf{k}}_{\mathbf{X}}(\boldsymbol{\theta}_t,t)$
and substituting into Eq.~\eqref{eq:cts-flow-kl} gives
\begin{equation}
\mathcal{L}_{\text{flow}}(\mathbf{X})
=
\frac{1}{2}
\int_0^1
\alpha(t)\;
\mathop{\mathbb{E}}\limits_{\boldsymbol{\theta}_t \sim p_F(\boldsymbol{\theta}\mid \mathbf{X};t)}
\big[
\|\mathbf{X}-\hat{\mathbf{k}}_{\mathbf{X}}(\boldsymbol{\theta}_t,t)\|_2^2
\big]\,\mathrm{d}t.
\label{eq:cts-flow-alpha}
\end{equation}

The channel accuracy is parameterized by a monotonically increasing function $\beta(t)$ with
$\beta(0)=0$ and $\beta(1)=\sigma_{1_\mathbf{X}}^{-2}$, and
\begin{equation}
\alpha(t) = \frac{\mathrm{d}}{\mathrm{d}t}\beta(t).
\end{equation}
We use
\begin{equation}
\beta(t) = \sigma_{1_\mathbf{X}}^{-2t}
\quad\Rightarrow\quad
\alpha(t)
=
- 2 \ln \sigma_{1_\mathbf{X}} \;\sigma_{1_\mathbf{X}}^{-2t}.
\end{equation}
Thus
\begin{equation}
\frac{1}{2}\alpha(t)
=
- \ln \sigma_{1_\mathbf{X}} \;\sigma_{1_\mathbf{X}}^{-2t},
\end{equation}
and Eq.~\eqref{eq:cts-flow-alpha} becomes
\begin{equation}
\mathcal{L}_{\text{flow}}(\mathbf{X}) = 
- \ln \sigma_{1_\mathbf{X}}
\int_0^1
\mathop{\mathbb{E}}\limits_{\boldsymbol{\theta}_t \sim p_F(\boldsymbol{\theta}\mid \mathbf{X};t)}
\left[
    \frac{\big\| \mathbf{X} - \hat{\mathbf{k}}_{\mathbf{X}}(\boldsymbol{\theta}_t,t) \big\|_2^2}
         {\sigma_{1_\mathbf{X}}^{2t}}
\right]\mathrm{d}t.
\end{equation}

Writing the integral as an expectation over $t \sim \mathcal{U}(0,1)$ yields the continuous-time loss
\begin{equation}
\mathcal{L}_{\infty}(\mathbf{X}) = 
- \ln \sigma_{1_\mathbf{X}}\;
\mathop{\mathbb{E}}\limits_{t\sim\mathcal{U}(0,1),\,
           \boldsymbol{\theta}_t \sim p_F(\boldsymbol{\theta}\mid \mathbf{X};t)}
\left[
\frac{\|\mathbf{X}-\hat{\mathbf{k}}_{\mathbf{X}}(\boldsymbol{\theta}_t,t)\|_2^2}
     {\sigma_{1_\mathbf{X}}^{2t}}
\right].
\end{equation}

The same derivation for the continuous bond representation $\mathbf{A}$ with its own target scale
$\sigma_{1_\mathbf{A}}$ gives $\mathcal{L}_{\infty}(\mathbf{A})$; the total training loss is
\[
\mathcal{L}_{\mathbf{G}} = \mathcal{L}_{\mathbf{X}} + \mathcal{L}_{\mathbf{A}}.
\]

\section{Proof of Theorem~\ref{thm:line_graph}}
\label{app:qw:edge_cost}

\begin{proof}
Let $e=(i,j)$ and $e'=(u,v)$ be two distinct bonds and denote
$D:=C^{A}_{e,e'}=\min_{p\in\{i,j\},\,q\in\{u,v\}}\mathrm{dist}_G(p,q)$.
\begin{align}
\mathrm{dist}_{\mathcal{L}(G)}(e,e')
&\le
\big|(e,(p_0,p_1),\ldots,(p_{D-1},p_D),e')\big|
\\
&=
D+1
=
1+C^{A}_{e,e'},
\end{align}
where $(p_0,\ldots,p_D)$ is a shortest $G$-path from an endpoint $p_0\in e$ to an endpoint $p_D\in e'$.
For any shortest $\mathcal{L}(G)$-path $(e=e_0,\ldots,e_L=e')$,
\begin{align}
C^{A}_{e,e'}
&=
\min_{p\in e,\,q\in e'}\mathrm{dist}_G(p,q)
\le
\mathrm{dist}_G(p_0,p_L)
\le
L-1
=
\mathrm{dist}_{\mathcal{L}(G)}(e,e')-1 .
\end{align}
Combining yields $\mathrm{dist}_{\mathcal{L}(G)}(e,e')=1+C^{A}_{e,e'}$.

For any coupling $\Pi$ with total mass $1$,
\begin{equation}
\langle \Pi,\mathrm{dist}_{\mathcal{L}(G)}\rangle
=
\langle \Pi,C^{A}\rangle + 1,
\end{equation}
so the additive shift does not change the optimal transport plan.
\end{proof}

\section{From Topology Matching to Class-Wise Transport Coupling}
\label{app:qw:derivation}

This appendix complements Sec.~\ref{sec:topo_ot} by making explicit the optimization principle behind Eqs.~\eqref{eq:qw_X} and~\eqref{eq:qw_A}. Rather than penalizing misclassification point-wise, we seek the minimum transport cost required to reshape predicted categorical mass into the ground-truth mass along the geodesic geometry of the molecular graph.

Let $P_X^{t}$ and $P_A^{t}$ denote the node/bond
categorical probabilities induced by $\boldsymbol{\theta}_t$ through Eq.~\eqref{eq:po-factorised},
and let $Q_X,Q_A$ be the one-hot targets derived from the ground-truth discrete graph.
We only couple classes that are present in the target to avoid ill-posed normalization:
\begin{align}
\mathcal{K}_X^{+}(G)=\Big\{k\in[K_X]:\sum_{i=1}^{N} Q_X(i,k) > 0\Big\}, \\
\mathcal{K}_A^{+}(G)=\Big\{k\in[K_A]:\sum_{e\in\mathcal{E}} Q_A(e,k) > 0\Big\},
\end{align}
and for bonds we exclude the no-bond class from $\mathcal{K}_A^{+}(G)$, as stated in Sec.~\ref{sec:topo_ot}.

Let $m_X$ and $m_A$ be the valid masks.
For each class $k$, we form class-wise mass fields on valid indices by simplex normalization:
\begin{align}
&\mathbf{a}_k^{X}=\mathrm{Norm}\!\big(P_X^{t}(:,k)\odot m_X\big),
\mathbf{b}_k^{X}=\mathrm{Norm}\!\big(Q_X(:,k)\odot m_X\big),
\\
&\mathbf{a}_k^{A}=\mathrm{Norm}\!\big(P_A^{t}(:,k)\odot m_A\big),
\mathbf{b}_k^{A}=\mathrm{Norm}\!\big(Q_A(:,k)\odot m_A\big),
\end{align}
where $\mathrm{Norm}(v)=v/\langle \mathbf{1},v\rangle$ whenever $\langle \mathbf{1},v\rangle>0$.

Given the geodesic ground costs $C^X$ and $C^A$ from Eq.~\eqref{eq:cost_X} and~\eqref{eq:cost_A}, we follow our topology-matching principle that
for each semantic class, transport the predicted mass to the target mass with minimum geodesic cost,
then aggregate these costs across classes. This yields Eqs.~\eqref{eq:qw_X} and~\eqref{eq:qw_A}.

\subsection{A single optimization view.} The class-wise aggregation admits a clean equivalence to one OT problem on a product space, which clarifies what is being optimized and why it is transport-coupled.

\begin{theorem}[Block-diagonal OT equivalence]
\label{thm:block_equiv}
Fix $C\in\mathbb{R}_+^{N \times N}$ and $\varepsilon>0$. Let $\{\mathbf{a}_k\}_{k=1}^{K},\{\mathbf{b}_k\}_{k=1}^{K}\subset \Delta^{N-1}$ be class-wise masses. Define measures on the product index set $[N]\times[K]$ by $\mu(i,k)=\frac{1}{K}\mathbf{a}_k(i)$ and $\nu(i,k)=\frac{1}{K}\mathbf{b}_k(i)$. Consider the cost on the product space:
\begin{equation}
\bar{C}\big((i,k),(j,k')\big)=
\begin{cases}
C_{ij}, & k=k',\\
+\infty, & k\neq k'.
\end{cases}
\end{equation}
Then the entropic OT objective on the product space decomposes as
\begin{equation}
\mathcal{W}_{\varepsilon}(\mu,\nu;\bar{C})
=
\frac{1}{K}\sum_{k=1}^{K}\mathcal{W}_{\varepsilon}(\mathbf{a}_k,\mathbf{b}_k;C).
\end{equation}
\end{theorem}

\begin{proof}
We prove the decomposition by two chains of inequalities.

\begin{align}
\mathcal{W}_{\varepsilon}(\mu,\nu;\bar C)
&=
\min_{\Pi\ge 0}\ \langle \Pi,\bar C\rangle
+\varepsilon\sum_{(i,k),(j,k')} \Pi_{(i,k),(j,k')}\big(\log \Pi_{(i,k),(j,k')}-1\big)
\label{eq:bd0}\\
&\text{s.t.}\quad
\sum_{j,k'} \Pi_{(i,k),(j,k')}=\mu(i,k),\qquad
\sum_{i,k} \Pi_{(i,k),(j,k')}=\nu(j,k').
\notag
\end{align}

\begin{align}
\mathcal{W}_{\varepsilon}(\mu,\nu;\bar C)
&\ge
\min_{\substack{\Pi\ge 0\\ \Pi_{(i,k),(j,k')}=0,\ k\neq k'}}\ 
\langle \Pi,\bar C\rangle
+\varepsilon\sum_{(i,k),(j,k')} \Pi_{(i,k),(j,k')}\big(\log \Pi_{(i,k),(j,k')}-1\big)
\label{eq:bd1}\\
&=
\min_{\{\Pi_k\}_{k=1}^{K}}\ 
\frac{1}{K}\sum_{k=1}^{K}
\Big(
\langle \Pi_k,C\rangle
+\varepsilon\sum_{i,j}(\Pi_k)_{ij}\big(\log (\Pi_k)_{ij}-1\big)
\Big)
-\varepsilon\log K
\label{eq:bd2}\\
&\text{s.t.}\quad
\Pi_k\ge 0,\ \Pi_k\mathbf{1}=\mathbf{a}_k,\ \Pi_k^\top\mathbf{1}=\mathbf{b}_k,\ \forall k,
\notag\\
&=
\frac{1}{K}\sum_{k=1}^{K}\mathcal{W}_{\varepsilon}(\mathbf{a}_k,\mathbf{b}_k;C)\;-\;\varepsilon\log K.
\label{eq:bd3}
\end{align}

\begin{align}
\mathcal{W}_{\varepsilon}(\mu,\nu;\bar C)
&\le
\langle \Pi^\star,\bar C\rangle
+\varepsilon\sum_{(i,k),(j,k')} \Pi^\star_{(i,k),(j,k')}\big(\log \Pi^\star_{(i,k),(j,k')}-1\big)
\label{eq:bd4}\\
&=
\frac{1}{K}\sum_{k=1}^{K}\mathcal{W}_{\varepsilon}(\mathbf{a}_k,\mathbf{b}_k;C)\;-\;\varepsilon\log K,
\label{eq:bd5}
\end{align}
where $\Pi^\star$ is constructed from the class-wise optimizers $\{\Pi_k^\star\}$ by
\[
\Pi^\star_{(i,k),(j,k')}=\frac{1}{K}(\Pi_k^\star)_{ij}\,\mathbf{1}[k=k'].
\]

Combining \eqref{eq:bd3}--\eqref{eq:bd5} yields
\begin{equation} 
\mathcal{W}_{\varepsilon}(\mu,\nu;\bar C)
=
\frac{1}{K}\sum_{k=1}^{K}\mathcal{W}_{\varepsilon}(\mathbf{a}_k,\mathbf{b}_k;C)\;-\;\varepsilon\log K.
\end{equation}
Since $-\varepsilon\log K$ is a constant independent of $\{\Pi_k\}$, it can be absorbed into the objective without affecting the optimal coupling. Therefore, up to an additive constant,
\begin{equation}
\mathcal{W}_{\varepsilon}(\mu,\nu;\bar C)
\equiv
\frac{1}{K}\sum_{k=1}^{K}\mathcal{W}_{\varepsilon}(\mathbf{a}_k,\mathbf{b}_k;C),
\end{equation}
which is the stated decomposition.
\end{proof}

\subsection{Proof of Theorem~\ref{thm:em}}
\label{proof:theorem_2}
\begin{proof}
Fix a class $k$ and consider unregularized OT
\begin{equation}
\mathcal{W}_{0}(\mathbf{a}_k,\mathbf{b}_k;C)
=
\min_{\Pi\ge 0}\ \langle \Pi, C\rangle
\quad
\text{s.t.}\quad
\Pi\mathbf{1}=\mathbf{a}_k,\ \Pi^\top\mathbf{1}=\mathbf{b}_k .
\label{eq:w0_def_em}
\end{equation}
Let $r=\sum_{i=1}^{N}Q(i,k)$ be the number of valid indices of class $k$ in the target, with $I=\{i_1,\ldots,i_r\}$ (predicted locations) and $J=\{j_1,\ldots,j_r\}$ (target locations). In the hard-label limit,
\begin{equation}
\mathbf{a}_k=\frac{1}{r}\sum_{\ell=1}^{r}\mathbf{e}_{i_\ell},
\qquad
\mathbf{b}_k=\frac{1}{r}\sum_{\ell=1}^{r}\mathbf{e}_{j_\ell}.
\label{eq:hard_masses_em}
\end{equation}
The marginal constraints imply $\Pi_{uv}=0$ unless $(u,v)\in I\times J$; define the restricted matrix $\widetilde{\Pi}\in\mathbb{R}_+^{r\times r}$ and cost $\widetilde{C}\in\mathbb{R}_+^{r\times r}$ by
\begin{equation}
\widetilde{\Pi}_{\ell m}=r\,\Pi_{i_\ell j_m},
\qquad
\widetilde{C}_{\ell m}=C_{i_\ell j_m}.
\label{eq:restrict_em}
\end{equation}
Then $\widetilde{\Pi}\mathbf{1}=\mathbf{1}$ and $\widetilde{\Pi}^\top\mathbf{1}=\mathbf{1}$, and
\begin{align}
\langle \Pi, C\rangle
&=
\sum_{\ell,m}\Pi_{i_\ell j_m}C_{i_\ell j_m} \\
&=
\frac{1}{r}\sum_{\ell,m}\widetilde{\Pi}_{\ell m}\widetilde{C}_{\ell m} \\
&=
\frac{1}{r}\langle \widetilde{\Pi},\widetilde{C}\rangle.
\label{eq:obj_reduce_em}
\end{align}
Therefore
\begin{equation}
\mathcal{W}_{0}(\mathbf{a}_k,\mathbf{b}_k;C)
=
\frac{1}{r}\min_{\widetilde{\Pi}\in\mathcal{B}_r}\ \langle \widetilde{\Pi},\widetilde{C}\rangle,
\qquad
\mathcal{B}_r=\{\widetilde{\Pi}\ge 0:\widetilde{\Pi}\mathbf{1}=\mathbf{1},\ \widetilde{\Pi}^\top\mathbf{1}=\mathbf{1}\}.
\label{eq:birkhoff_em}
\end{equation}
Since the objective is linear, an optimum is attained at an extreme point of $\mathcal{B}_r$. By Birkhoff--von Neumann~\cite{}, the extreme points are permutation matrices $P_\pi$ ($\pi\in\mathfrak{S}_r$), hence
\begin{equation}
\mathcal{W}_{0}(\mathbf{a}_k,\mathbf{b}_k;C)
=
\frac{1}{r}\min_{\pi\in\mathfrak{S}_r}\sum_{\ell=1}^{r} C_{i_\ell, j_{\pi(\ell)}}.
\label{eq:assignment_em}
\end{equation}
Thus, in the hard-label regime, class-wise OT reduces to within-class minimum-cost matching under the geodesic cost $C$, making long-range permutations strictly more expensive than local corrections.
\end{proof}

\begin{table*}
  \centering
  \renewcommand \arraystretch{1.0}
  \begin{tabular}{cccc}
    \toprule
    Category & Hyper-parameter & Description & Value \\
    \midrule
    
    \multirow{4}{*}{Bayesian Flow} & $\sigma_{1, \mathbf{X}}$ & The standard deviation of $\mathbf{X}$ of the noise distribution & 0.2 \\
    & $\sigma_{1, \mathbf{A}}$ & The standard deviation of $\mathbf{A}$ of the noise distribution & 0.2 \\
    & $T$ & Sampling steps & 200 \\
    & $t_{min}$ & Minimum time & $10^{-4}$ \\

    \midrule
    \multirow{3}{*}{QW Transport} & $\lambda_{QW}$ & The coefficient weighting the QW structural regularizer & 0.1 \\
    & $\epsilon_{QW}$ & The entropic regularization smoothing the optimal transport plan calculation & 0.2 \\
    & $n_{QW}$ & The number of Sinkhorn iterations of the entropic Optimal Transport problem & 30 \\
    
    \bottomrule
  \end{tabular}
  \caption{Hyper-parameters of TopBF.}
  \label{tab:hp}
\end{table*}

\section{Experimental Details}

\subsection{Datasets}

We validate our approach on the QM9 and ZINC250k benchmarks, following the previous experimental setup~\cite{jo2022score}. QM9 contains approximately 133,885 molecules with up to 9 heavy atoms of 4 types, while ZINC250k comprises roughly 249,455 molecules with up to 38 heavy atoms of 9 types, both covering three standard bond types. Adhering to the standard pipeline~\cite{shi2020graphaf,luo2021graphdf,jo2022score}, molecules undergo kekulization via RDKit, with explicit hydrogens excluded to focus on the heavy-atom graph representation.

\subsection{Implementation details}

For our TopBF, we train our model with a constant learning rate $10^{-4}$ with AdamW optimizer and weight decay $10^{-12}$, applying the exponential moving average (EMA) to the parameters. The core network $\Psi(\cdot)$ of TopBF is the graph transformer~\cite{dwivedi2020generalization,vignac2023digress}, with the inner parameters consistent across all baselines using it. Other critical hyper-parameters are presented in Table~\ref{tab:hp}.

\section{Supplementary Experiments}

\subsection{Computational Efficiency and Wall-Clock Time Analysis}
\label{app:efficiency}

We conduct a rigorous wall-clock time comparison against state-of-the-art baselines. We repeat the measurement and report the average.

\paragraph{Experimental Setup}
To ensure a fair comparison, all runtime measurements were conducted under a strictly controlled hardware and software environment. 
\begin{itemize}
    \item \textbf{Hardware:} All models were trained and evaluated on a single NVIDIA A6000 GPU with an AMD EPYC 7763 (64 cores) @ 2.450GHz CPU.
    \item \textbf{Software:} Experiments were implemented in PyTorch 2.0.1 with CUDA 11.7.
    \item \textbf{Controls:} We fixed the batch size to 1,024 for QM9 and 256 for ZINC250. For sampling efficiency, we measured the total time required to generate 10,000 valid molecules with their own reported sampling steps.
\end{itemize}

We compare TopBF against three representative baselines: GDSS~\cite{jo2022score}, DiGress~\cite{vignac2023digress}, and SID~\cite{boget2025simple}. These methods represent the current standard for score-based and discrete generative models, respectively.

\begin{table}[h]
\centering
\label{tab:training_time}
\begin{tabular}{c|cc}
\toprule
Method & QM9 (min)$\downarrow$ & ZINC250k (min)$\downarrow$ \\
\midrule
GDSS    & 1.42 & 26.47 \\
DiGress & 10.08 & 212.95 \\
SID & 1.58 & 25.78 \\
\textbf{TopBF (Ours)} & \textbf{1.33} & \textbf{23.32} \\
\bottomrule
\end{tabular}
\caption{Comparison of total sampling time. Lower is better.}
\label{tab:wall_clock}
\end{table}

As shown in Table~\ref{tab:wall_clock}, TopBF maintains its lead by generating 10,000 valid molecules in just 1.33 minutes for QM9 and 23.32 minutes for ZINC250k. 
These results indicate that TopBF achieves competitive or improved sampling throughput under our controlled setup. Since QW is used only as a training-time regularizer, it does not introduce additional cost during sampling.

\begin{figure}
    \centering
    \includegraphics[width=.95\linewidth]{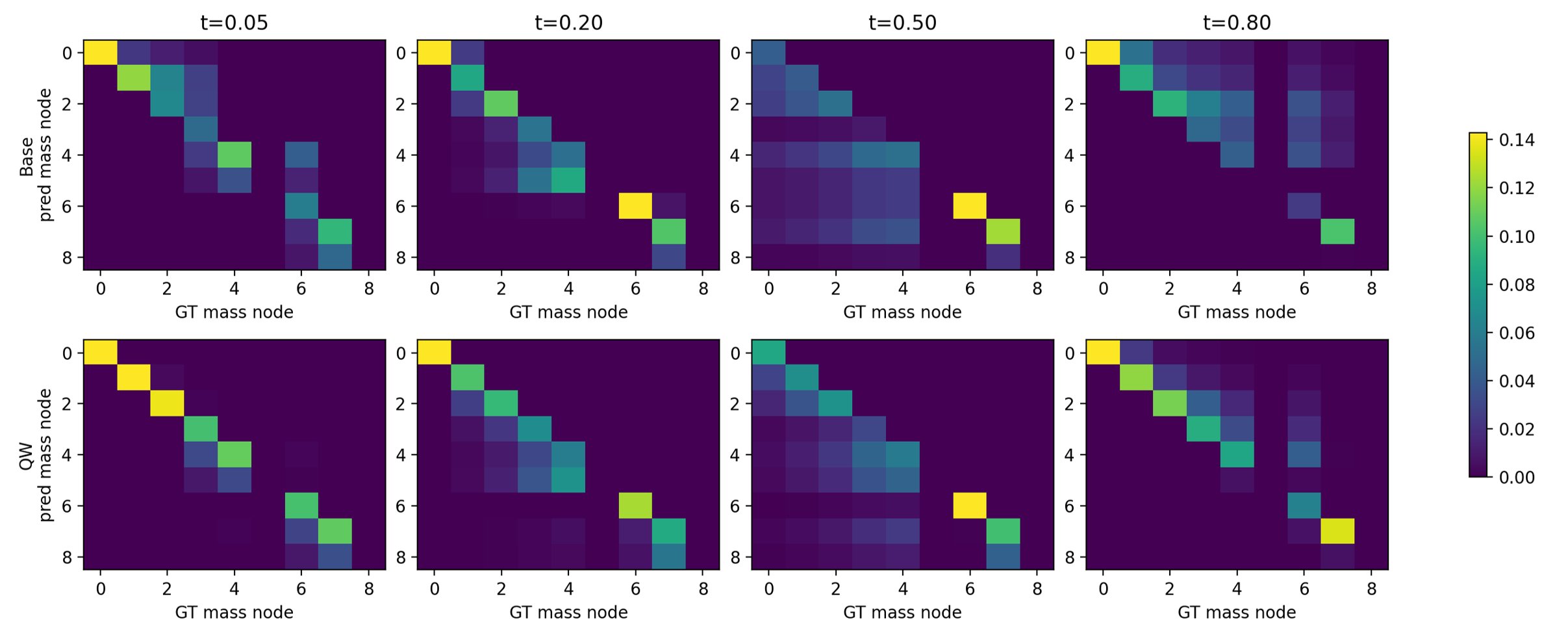}
    \caption{Sinkhorn coupling $P$ between the predicted node-type marginals and the ground truth under the shortest-path (geodesic) cost, probed at $t\in\{0.05,0.2,0.5,0.8\}$ on QM9 dataset.}
    \label{fig:ot-coupling-qm9}
\end{figure}

\begin{table}[t]
\centering
\renewcommand\arraystretch{1.15}
\begin{tabular}{c|ccc|ccc|ccc|ccc}
\toprule
\multirow{2}{*}{$t$} &
\multicolumn{3}{c|}{cost\_avg$\downarrow$} &
\multicolumn{3}{c|}{localmass1$\uparrow$} &
\multicolumn{3}{c|}{localmass2$\uparrow$} &
\multicolumn{3}{c}{avg\_pX\_gt$\uparrow$} \\
\cmidrule(lr){2-4}\cmidrule(lr){5-7}\cmidrule(lr){8-10}\cmidrule(lr){11-13}
& Base & QW & Improve & Base & QW & Improve & Base & QW & Improve & Base & QW & Improve \\
\midrule
0.05 & 0.834 & 0.605 & \textbf{27.5\%} & 0.810 & 0.852 & \textbf{5.2\%} & 0.911 & 0.923 & \textbf{1.3\%} & 0.328 & 0.495 & \textbf{50.9\%} \\
0.20 & 0.815 & 0.767 & \textbf{5.9\%}  & 0.800 & 0.813 & \textbf{1.6\%} & 0.901 & 0.911 & \textbf{1.1\%} & 0.347 & 0.602 & \textbf{73.5\%} \\
0.50 & 1.454 & 1.149 & \textbf{21.0\%} & 0.612 & 0.698 & \textbf{14.1\%} & 0.755 & 0.826 & \textbf{9.4\%} & 0.422 & 0.581 & \textbf{37.7\%} \\
0.80 & 0.800 & 0.344 & \textbf{57.0\%} & 0.761 & 0.913 & \textbf{20.0\%} & 0.913 & 0.987 & \textbf{8.1\%} & 0.552 & 0.707 & \textbf{28.1\%} \\
\bottomrule
\end{tabular}
\caption{Probing statistics under shortest-path cost at $t\in\{0.05,0.2,0.5,0.8\}$ (Base vs.\ QW). QW reduces transport cost and increases local transport mass, indicating more geodesically local corrections.}
\label{tab:probing_short}
\end{table}


\subsection{Interpretability Analysis}
\label{app:interpretability}

QW regularization is intended to encourage topology-consistent (geodesically local) corrections. To make this effect observable beyond final metrics, we report controlled \emph{probing} at fixed noise levels $t$. We evaluate four representative noise levels $t\in\{0.05,0.2,0.5,0.8\}$.

For probing, we quantify locality via Sinkhorn-OT diagnostics under shortest-path cost: transport cost \texttt{cost\_avg}$\downarrow$ and local-mass ratios within 1/2 hops \texttt{localmass1/2}$\uparrow$, together with a correctness proxy \texttt{avg\_pX\_gt}$\uparrow$.
Improve is relative improvement: for $\downarrow$, $(\text{Base}-\text{QW})/\text{Base}$; for $\uparrow$, $(\text{QW}-\text{Base})/\text{Base}$.

Across noise levels, probing verifies the intended mechanism: QW promotes geodesically local corrections (lower $\texttt{cost\_avg}$, higher $\texttt{localmass1/2}$).

\begin{figure}
    \centering
    \includegraphics[width=.95\linewidth]{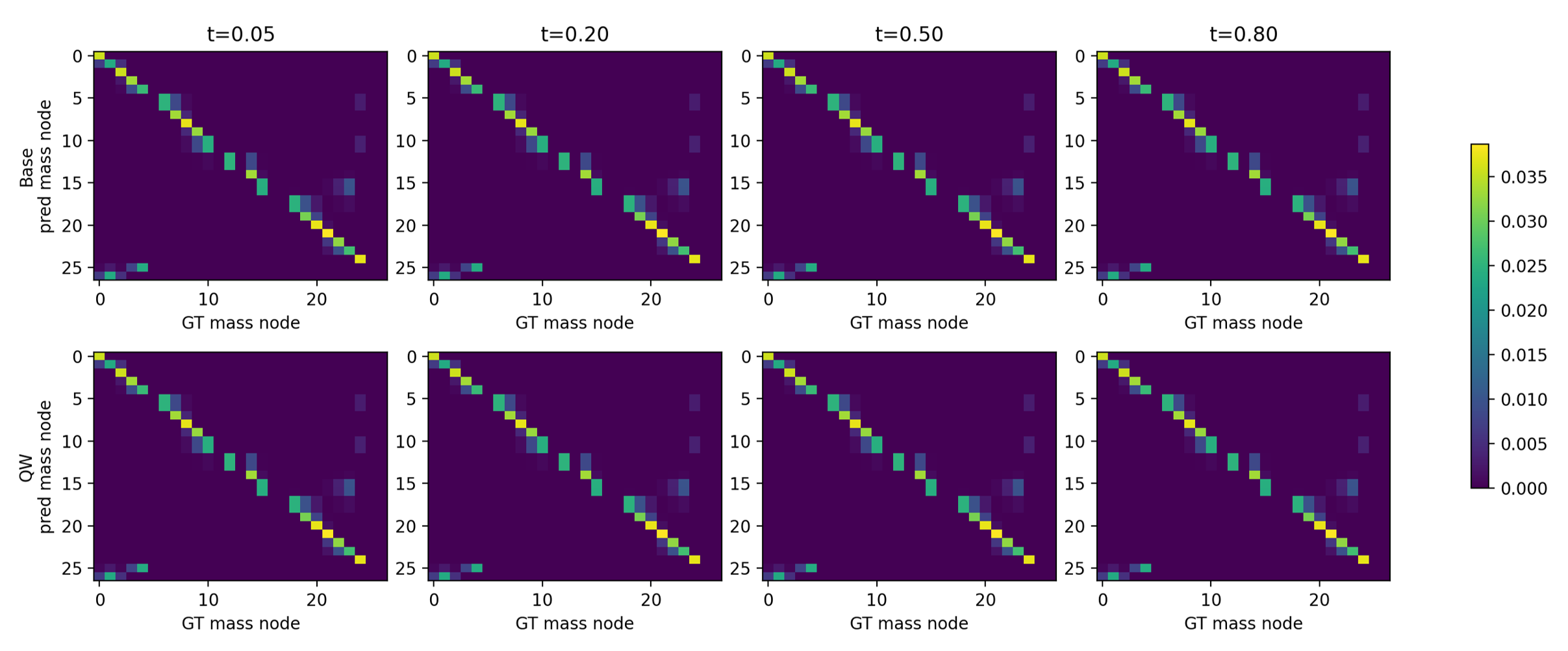}
    \caption{Sinkhorn coupling $P$ between the predicted node-type marginals and the ground truth under the shortest-path (geodesic) cost, probed at $t\in\{0.05,0.2,0.5,0.8\}$ on ZINC250k dataset.}
    \label{fig:ot-coupling-zinc250k}
\end{figure}

The coupling matrix $P$ transports the predicted node-type mass to the ground-truth mass with graph geodesic distance as the unit cost.
Mass concentrated near small distances indicates that correcting node-type distributions primarily involves local adjustments on the graph, whereas diffuse off-diagonal mass implies long-range relocation.
As shown in Figures~\ref{fig:ot-coupling-qm9} and~\ref{fig:ot-coupling-zinc250k}, QW produces a noticeably more localized transport plan than the Base model at the same noise level, consistent with the reduced \texttt{cost\_avg} and increased local-mass ratios in Table~\ref{tab:probing_short}.

\section{Visualization of Generated Molecular Graphs}
\label{app:vis}

\begin{figure*}[!htbp]
    \centering
    \includegraphics[width=0.95\linewidth]{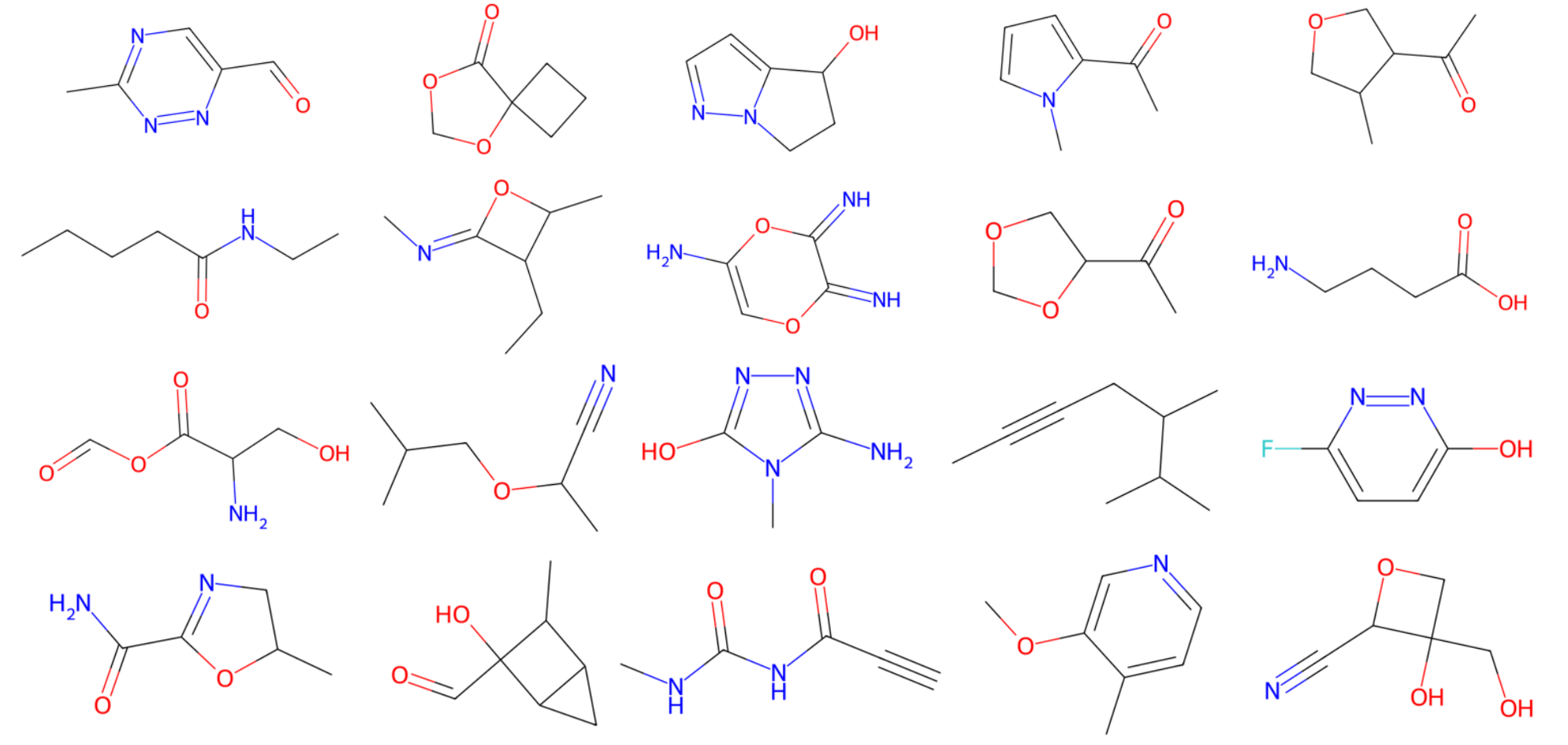}
    \caption{Molecular graphs generated by TopBF on QM9 dataset.}
    \label{fig:samples_q}
\end{figure*}

\begin{figure*}[!htbp]
    \centering
    \includegraphics[width=0.95\linewidth]{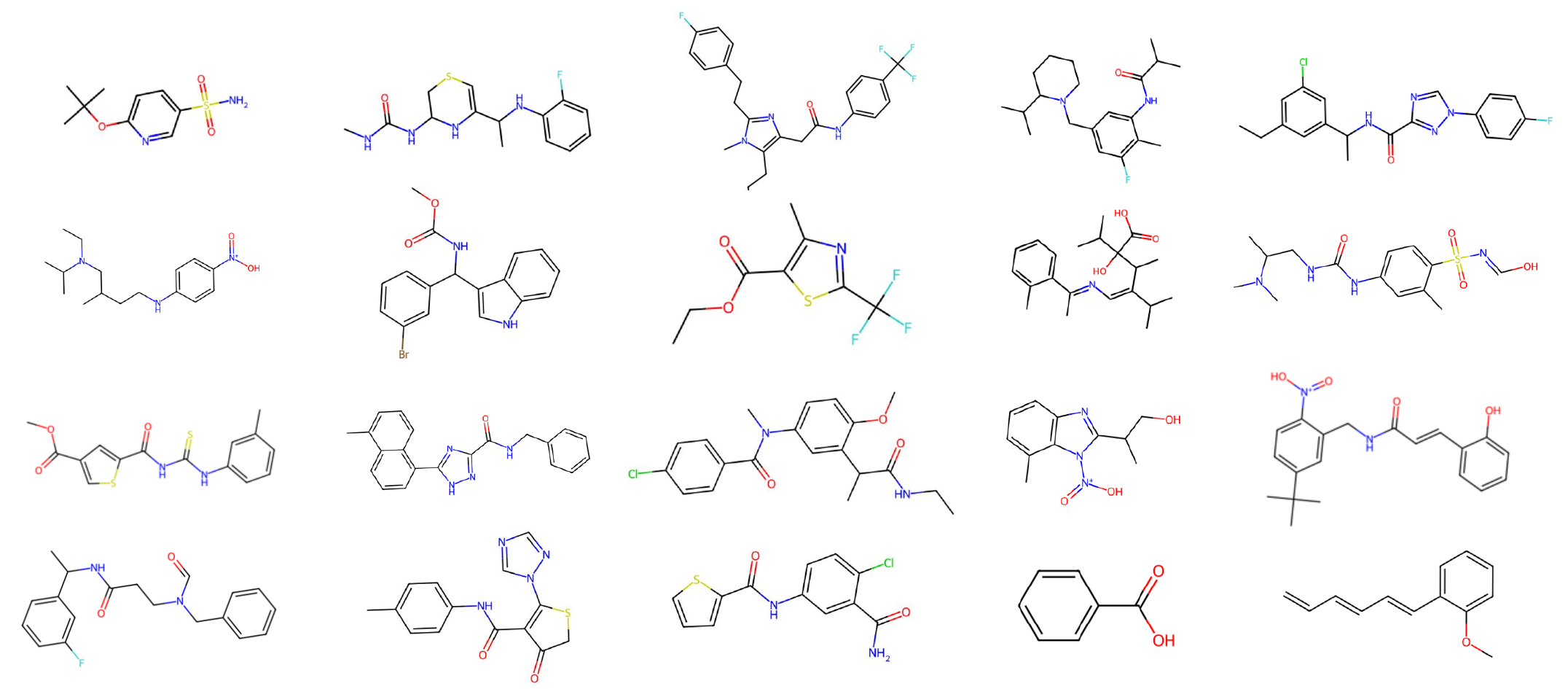}
    \caption{Molecular graphs generated by TopBF on ZINC250k dataset.}
    \label{fig:samples_z}
\end{figure*}

We demonstrate some randomly selected molecular graphs generated by TopBF on both QM9 and ZINC250k datasets in Figures.~\ref{fig:samples_q} and~\ref{fig:samples_z}.

\section{Pseudocode for TopBF}
\label{app:pseudocode}

Pseudocode for essential functions in TopBF and continuous-time loss $\mathcal{L}_{\infty}$ are presented in Algorithms~\ref{alg:funtions} and~\ref{alg:train}, while sampling procedure is provided in Algorithm~\ref{alg:sample}, wherein $\text{NEAREST\_CENTER}$ compares the inputs to the nearest centers. 

\clearpage
\begin{multicols}{2} 

\begin{algorithm}[H]
\caption{Functions for TopBF}
\label{alg:funtions}
\begin{algorithmic}

\STATE \textbf{function} \textsc{CDF}$(\mu, \sigma, x)$
    \STATE $F(x) \gets \dfrac{1}{2}\left[1 + \mathrm{erf}\bigl(\dfrac{x-\mu}{\sqrt{2}\sigma}\bigr)\right]$
    \IF{$x \le -1$}
        \STATE $H(x) \gets 0$
    \ELSIF{$x \ge 1$}
        \STATE $H(x) \gets 1$
    \ELSE
        \STATE $H(x) \gets F(x)$
    \ENDIF
    \STATE \textbf{return} $H(x)$
\STATE \textbf{end function}

\STATE \textbf{function} \text{PRED}$(\mathbf{\mu} \in \mathbb{R}^{D\times K},\, t \in [0,1],\, K \in \mathbb{N},\, \gamma(t) \in \mathbb{R}^+,\, t_{\min} \in \mathbb{R}^+)$
    \STATE \# $t_{\min}$ is set to $0.0001$ by default
    \IF{$t < t_{\min}$}
        \STATE $\hat{\mu} \gets 0,\,\, \hat{\sigma} \gets 1$
    \ELSE
        \STATE $\boldsymbol{\mu}_\epsilon, \ln \boldsymbol{\sigma}_\epsilon \gets \Psi(\boldsymbol{\theta}_t, \mathbf{c}, t)$
        \STATE $\hat{\boldsymbol{\mu}} \gets \frac{\boldsymbol{\mu}}{\gamma(t)} - \sqrt{\frac{1-\gamma(t)}{\gamma(t)}}\boldsymbol{\mu}_\epsilon$
        \STATE $\hat{\boldsymbol{\sigma}} \gets
            \sqrt{\frac{1-\gamma(t)}{\gamma(t)}}\text{exp}(\ln\boldsymbol{\sigma}_\epsilon)$
    \ENDIF

    \FOR{$d = 1,\dots,D$}
        \FOR{each $k \in K$}
            \STATE $p_O^{(d)}(k\,|\,\theta) \gets
                \text{CDF}(\hat{\mu}^{(d)},\hat{\sigma}^{(d)},k_r)
                - \text{CDF}(\hat{\mu}^{(d)},\hat{\sigma}^{(d)},k_l)$
        \ENDFOR
    \ENDFOR

    \STATE \textbf{return} $p_O(\cdot \mid \boldsymbol{\theta}_t)$
\STATE \textbf{end function}

\end{algorithmic}
\end{algorithm}


\begin{algorithm}[H]
    \caption{Training Algorithm} 
    \begin{algorithmic}
        \REQUIRE $\sigma_{1}\in \mathbb{R}^+, K \in \mathbb{N}$
        \STATE \textbf{Input:} parameterized continuous graph $\mathbf{G}$
        \STATE $t \sim \mathcal{U}(0, 1)$
        \STATE $\gamma \gets 1 - \sigma_{1}^{2t}$
        \STATE $\mathbf{\mu} \sim \mathcal{N}(\gamma(t), \gamma(t)(1-\gamma(t))\boldsymbol{I})$
        \STATE $p_{O}(\cdot \mid \boldsymbol{\theta}_t) \gets \text{PRED}(\boldsymbol{\mu}, t, K, \gamma(t))$
        \STATE $\hat{\mathbf{k}}_{c}(\boldsymbol{\theta}_t, t) \gets \text{NEAREST\_CENTER}\big(p_O(\cdot \mid \boldsymbol{\theta}_t)\big)$
        \STATE $\mathcal{L}_{\infty}(\mathbf{G}) \gets -\ln \sigma_{1}\sigma_{1}^{-2t}\|\mathbf{G}-\hat{\mathbf{k}}_{c}(\boldsymbol{\theta}_t, t)\|_2^2$ 
        \RETURN $\mathcal{L}_{\infty}(\mathbf{G})$
    \end{algorithmic} 
    \label{alg:train}
\end{algorithm}


\begin{algorithm}[H]
    \caption{Sampling Algorithm} 
    \begin{algorithmic}
        \REQUIRE $\sigma_{1}\in \mathbb{R}^+$, number of steps $n$, $K \in \mathbb{N}$, optional $\mathbf{z}^\star$, $\lambda \geq 0$
        \STATE $\boldsymbol{\mu}_0 \gets \mathbf{0}, \,\,\rho_0 \gets 1$
        \FOR{$i=1\;\textbf{to}\;n$}
            \STATE $t \gets \frac{i-1}{n}$
            \STATE $\gamma(t) \gets 1-\sigma_{1}^{2t}$
            \STATE $\boldsymbol{\theta}_t \gets (\boldsymbol{\mu}_{i-1}, \rho_{i-1})$
            \IF{$\mathbf{z}^\star$ is not None \textbf{and} $\lambda > 0$}
                \STATE $\hat{\mathbf{z}}_t \gets R_\varphi(\boldsymbol{\theta}_t, t)$
                \STATE $\mathcal{L}_{\text{prop}} \gets \|\hat{\mathbf{z}}_t - \mathbf{z}^\star\|_2^2$
                \STATE $\boldsymbol{\mu}_{i-1} \gets \boldsymbol{\mu}_{i-1}
                    - \lambda \nabla_{\boldsymbol{\mu}_{i-1}} \mathcal{L}_{\text{prop}}$
                \STATE $\boldsymbol{\theta}_t \gets (\boldsymbol{\mu}_{i-1}, \rho_{i-1})$
            \ENDIF
            \STATE $p_{O}(\cdot \mid \boldsymbol{\theta}_t) \gets \text{PRED}(\boldsymbol{\mu}_{i-1}, t, K, \gamma(t))$
            \STATE $\alpha(t) \gets \sigma_{1}^{-2i/n}\big( 1 - \sigma_{1}^{2/n} \big)$
            \STATE $\mathbf{k}_{c}(\boldsymbol{\theta}_t, t) \gets \text{NEAREST\_CENTER}\big(p_O(\cdot \mid \boldsymbol{\theta}_t)\big)$
            \STATE $\mathbf{Y} \sim \mathcal{N}(\mathbf{k}_{c}(\boldsymbol{\theta}_t, t), \alpha(t)^{-1}\boldsymbol{I})$
            \STATE $\boldsymbol{\mu}_{i} \gets \dfrac{\rho_{i-1} \boldsymbol{\mu}_{i-1} + \alpha(t) \mathbf{Y}}{\rho_{i-1} + \alpha(t)}$
            \STATE $\rho_{i} \gets \rho_{i-1} + \alpha(t)$
        \ENDFOR
        \STATE $p_{O}(\cdot \mid \boldsymbol{\theta}_1) \gets \text{PRED}(\boldsymbol{\mu}_{n}, 1, K, 1-\sigma_{1}^2)$
        \STATE $k'_{c}(\boldsymbol{\theta}_1, 1) \gets \text{NEAREST\_CENTER}\big(p_O(\cdot \mid \boldsymbol{\theta}_1)\big)$
        \RETURN $k'_{c}(\boldsymbol{\theta}_1, 1)$
    \end{algorithmic} 
    \label{alg:sample}
\end{algorithm}
\end{multicols}



\end{document}